\documentclass[review,authoryear]{elsarticle}
\usepackage[top=1in, bottom=1in, left=1.25in, right=1.25in]{geometry}
\usepackage{times}
\usepackage{amssymb}
\usepackage{graphicx}
\usepackage{amsmath}
\usepackage{mathrsfs}
\usepackage{epstopdf}
\usepackage{multirow}
\usepackage{subcaption} 
\usepackage{kotex}
\usepackage{tikz}
\usetikzlibrary{matrix}
\usetikzlibrary{positioning}
\usepackage{color}
\usepackage{cite}
\usepackage{pifont}
\usepackage{url}
\usepackage{hyperref}
\usepackage{caption}
\usepackage{sidecap}
\usepackage{rotating}
\usepackage{chngcntr}

\usepackage[flushleft]{threeparttable}
\usepackage{booktabs}
\usetikzlibrary{arrows,decorations.pathmorphing,decorations.markings,backgrounds,positioning,fit,plotmarks}
\usepackage{pgfplots}
\newcommand{\ie}{\textit{i.e.}}
\newcommand{\eg}{\textit{e.g.}}

\definecolor{red}{RGB}{255,0,0}
\definecolor{blue}{RGB}{0, 0, 255}
\definecolor{ballblue}{rgb}{0.13, 0.67, 0.8}

\newcommand{\quotes}[1]{``#1''}
\newcommand*{\rom}[1]{\expandafter\@slowromancap\romannumeral #1@}
\newcommand{\eat}[1]{}
\usepackage{url}

\journal{Medical Image Analysis}

\begin{document}
\begin{frontmatter}



\title{Deep Joint Learning of Pathological Region Localization and \\ Alzheimer's Disease Diagnosis}

\author[ku1]{Changhyun Park}
\author[ku1,ku2]{Heung-Il Suk\corref{cor1}}
\author{\\and the Alzheimer's Disease Neuroimaging Initiative\fnref{dataset}}
\address[ku1]{Department of Brain and Cognitive Engineering, Korea University, Seoul 02841, Republic of Korea}
\address[ku2]{Department of Artificial Intelligence, Korea University, Seoul 02841, Republic of Korea}
\cortext[cor1]{Corresponding author: Heung-Il Suk (hisuk@korea.ac.kr)}

\fntext[dataset]{Data used in preparation of this article were obtained from the Alzheimer's Disease Neuroimaging Initiative (ADNI) database (\url{http://www.loni.ucla.edu/ADNI}). As such, the investigators within the ADNI contributed to the design and implementation of ADNI and/or provided data but could not participate in analysis or writing of this report. A complete listing of ADNI investigators can be found at \url{http://adni.loni.ucla.edu/wpcontent/uploads/how_to_apply/ADNI_Authorship_List.pdf}.
}
\begin{abstract} 
The identification of Alzheimer's disease (AD) and its early stages using structural magnetic resonance imaging (MRI) has been attracting the attention of researchers. Various data-driven approaches have been introduced to capture subtle and local morphological changes of the brain accompanied by the disease progression. One of the typical approaches for capturing subtle changes is patch-level feature representation. However, the predetermined regions to extract patches can limit classification performance by interrupting the exploration of potential biomarkers. In addition, the existing patch-level analyses have difficulty explaining their decision-making because the final decision is made using nonlinear interactions of patch-level feature representations. To address these problems, we propose the BrainBagNet with a position-based gate (PG-BrainBagNet), a framework for jointly learning pathological region localization and AD diagnosis in an end-to-end manner. In advance, as all scans are aligned to a template in image processing, the position of brain images can be represented through the 3D Cartesian space shared by the overall MRI scans. Therefore, we represent coordinates in the 3D Cartesian space containing the volumetric position information of MRI scans by combining positions from the coronal, sagittal, and axial planes. Then, the proposed method represents the patch-level response from whole-brain MRI scans and discriminative brain-region from position information. Based on the outcomes, the patch-level class evidence is calculated, and then the image-level prediction is inferred by a transparent aggregation. The proposed models were evaluated on the Alzheimer's Disease Neuroimaging Initiative (ADNI) datasets (\ie, ADNI-1 and ADNI-2). In five-fold cross-validation, the classification performance of the proposed method outperformed that of the state-of-the-art methods in both AD diagnosis (AD vs. normal control) and mild cognitive impairment (MCI) conversion prediction (progressive MCI vs.  stable MCI) tasks. In addition, changes in the identified discriminant regions and patch-level class evidence according to the patch size used for model training are presented and analyzed. Code is available at: \url{github.com/ku-milab/PG-BrainBagNet}.
\end{abstract}

\end{frontmatter}

\section{Introduction}
\label{sec:introduction}
Alzheimer’s disease (AD) is a type of neurodegenerative brain disorder characterized by an irreversible and progressive loss of neurons \citep{jagust2013vulnerable}.
This degenerative disorder is also considered the most common cause of dementia \citep{barker2002relative}, which is commonly accompanied by such symptoms as memory loss and progression toward long-term impairment of cognitive functioning.
However, these symptoms are not manifest in the early stages of AD, and the symptoms gradually worsen without being recognized as the condition \citep{larsen2019data}.
Moreover, AD primarily progresses through a prodromal stage referred to as mild cognitive impairment (MCI).
Thus, in the past few decades, numerous studies \citep{mosconi2007early, gray2013random, liu2014hierarchical, rathore2017review, arbabshirani2017single, JUNG2021118143} have focused on both accurately classifying the AD group from the normal control (NC) group and predicting the MCI conversion/transition to detect the early stages of AD.
Typically, the MCI conversion prediction task is to distinguish between stable MCI (sMCI) and progressive MCI (pMCI) based on the risk of AD progression.
These studies on AD diagnosis and its early detection can help identify high-risk cohorts for better treatment planning and to further improve the quality of life \citep{fung2019alzheimer}.

Although the symptoms that appear in AD are not manifested in the early stages, the biological processes underlying the disease are present for decades before the symptoms occur \citep{bennett2006neuropathology, jagust2018imaging}.
Therefore, neuroimaging data, such as magnetic resonance imaging (MRI), have been employed for examining neurodegeneration, diagnosing AD, and detecting its early stages \citep{frisoni2010clinical, wolz2011multi, coupe2012scoring}.
Furthermore, brain atrophy is the most proximate substrate of cognitive impairment in AD; thus, automated computer-aided diagnosis based on structural MRI (sMRI) has been attracting attention as promising studies \citep{vemuri2010role, pmlr-v116-liu20a, tanveer2020machine}.
In addition, explaining diagnostic results as clinical evidence is crucial for clinical application \citep{lee2019toward, EITEL2019102003}.
Moreover, convolutional neural networks (CNNs) exhibited superior performance in image recognition and are effectively applied in various domains, including the medical field \citep{li2014medical, shin2016deep, korolev2017residual, brendel2018bagnets, schlemper2019attention}.
Likewise, three-dimensional (3D) CNN-based models can filter out task-oriented structural patterns such as brain atrophy from 3D sMRI scans.
Therefore, recent studies on AD diagnosis \citep{basaia2019automated, jin2020generalizable, pmlr-v116-liu20a, lian2020attention} have concentrated on computer-aided diagnosis systems based on sMRI scans with 3D CNNs.

Most existing 3D CNN-based AD diagnostic models \citep{korolev2017residual, fung2019alzheimer, jin2019attention, wang2020ensemble, pmlr-v116-liu20a, liu2018landmark, lian2018hierarchical, lian2020attention} can be categorized into two classes according to the input type.
One option is to take a 3D whole-brain image as input.
The other option is to take a bag of instances in the multiple instance learning (MIL) terminology, where each 3D whole brain image is regarded as a bag, and the 3D patches extracted from the bag are treated as instances.
When models take whole-brain images as input, features are hierarchically extracted, from local to global patterns, for an accurate AD diagnosis.
The models gradually expand the region size that produces the features (\ie, receptive field), and the size eventually becomes comparable to the input image size.
In this process, local and global structural patterns are captured and employed for making decisions.

Recent studies taking this practical feature representation approach put their efforts into better feature extraction by proposing novel network architectures.
For instance, as the AD progression accompanies local and subtle brain changes, \citep{pmlr-v116-liu20a} proposed a network architecture avoiding the early spatial downsampling to design architectures capable of learning subtle differences.
In addition, \citep{jin2019attention, jin2020generalizable} introduced an attention-based CNN model to automatically generate more discriminative feature representation in brain images.
However, deep learning methods are criticized for the difficulty of retracing the classification decision due to the vast parameter spaces and nonlinear interactions \citep{castelvecchi2016can}. 
In particular, in medical applications, interpretation of the decision-making (\eg, which brain regions contribute to decision-making) is essential for clinical integration, error tracking, and knowledge discovery \citep{EITEL2019102003}.
Although existing models \citep{bohle2019layer, EITEL2019102003, pmlr-v116-liu20a} have mitigated this problem by relying on additional applications that generate heatmaps for each subject, the voxel-wise relevance represented by heatmaps does not allow making a statement about the underlying reasons such as brain atrophy or pathological structural changes \citep{bohle2019layer}.

In the case of taking a bag of instances as input, various approaches \citep{LIU20121106, suk2014hierarchical, tong2014multiple} for patch-level feature representation have attempted to more efficiently characterize local structural changes induced by AD.
In addition, feature representation using the CNN model \citep{liu2018landmark, lian2018hierarchical, lian2020attention} has recently been proposed.
These approaches are intended to extract local and subtle structural changes by allowing models to extract features from the specific receptive field size with an upper bound on the patch size.
Conventional methods \citep{coupe2012scoring, 7929275} generally assign the class label of an image to all patches extracted from the corresponding image to extract patch-level feature representation.
However, the class labels for each patch are ambiguous because not all patches extracted from patients' brains include changes associated with pathology \citep{tong2013multiple}.
Even, the proportion of AD-induced brain atrophy might vary according to the individual patient.
In this circumstance, weakly supervised learning strategies, such as MIL, have been adopted \citep{tong2014multiple, liu2018landmark}. 
A given brain MRI scan is considered a bag, and the 3D patches that comprise the bag are treated as instances.
The simplest method to extract the patches is to use all of the patches over the brain. 
However, the number of patches that can be extracted from 3D high-resolution brain images is too large.
In addition, including numerous patches not associated with AD can lead to a low proportion of instances containing AD pathological changes in bags labeled as the AD class.
This method can cause serious class imbalance problems and degrade performance for many real-world problems \citep{carbonneau2016witness, carbonneau2018multiple}.
Therefore, localizing the pathological brain regions and extracting patches from these regions are an important and challenging problems.

For patch extraction, existing studies have introduced data-driven approaches \citep{LIU20121106, tong2014multiple, liu2018landmark, liu2018anatomical}, which commonly follow a standard pipeline. 
First, a discriminative probability map is generated by using group differences of local features.
Then, based on the probability map, patches are extracted by various hyperparameters, such as patch size, number of patches, and the minimum distance between patches.
These approaches have successfully extracted class-discriminative patches and presented an informative result, leading to a high increase in the classification performance of the diagnostic model.
However, the predetermined area to extract patches can lead to suboptimal classification performance because it has been independently performed without considering feature extraction and classification.
In addition, hyperparameters for patch extraction eventually cause time-consuming hyperparameter exploration or optimization procedures by iterating the diagnostic model training according to numerous hyperparameter combinations.

To alleviate issues arising from the independence of patch extraction and diagnostic model training, a hierarchical fully convolutional network (H-FCN) was proposed for joint atrophy localization and disease identification \citep{lian2018hierarchical}.
Hybrid loss was designed by gathering patch-, region-, and image-level loss for diagnostic model training and ranking the discriminative capacity of the corresponding location.
However, hybrid loss includes an objective such that the patch- and region-level classification scores become closer to the image-level ground-truth class labels, and this trial might include faulty assumptions because not all patches and regions might have been affected by AD.
In addition, the pruning approach cannot consider potential patch-level biomarkers not previously extracted as candidates.
More recently, a hybrid network (HybNet) \citep{lian2020attention} improved the H-FCN by fusing two branches, which could extract local and global structural information, respectively.
Although a novel discriminative brain-region localization approach has been introduced for patch extraction, the fundamental problem caused by brain-region predetermination remains.
In addition, as the proposed learning pipeline becomes increasingly complex, there are limitations regarding a lack of explanation of the prediction results despite competitive performance.

The current methods for patch-level feature representation have evolved from various perspectives based on the predetermined brain-region localization.
However, the dependency between discriminative brain-region localization, patch extraction, and the diagnostic model has usually been overlooked.
For instance, if diagnostic model predictions were made using 3D patches as large as a whole-brain image, most patches could contain pathological changes wherever the patches were extracted.
However, as 3D patches become smaller, the learning model can efficiently characterize subtle brain changes, whereas patches with pathological changes can only be detected in sparse areas of the brain.
Therefore, predetermination of a brain region to extract patches may limit the opportunity to explore potential biomarkers or lead to a low proportion of instances containing AD pathological changes in bags labeled as the AD class.
A joint learning framework for discriminative brain-region localization and disease diagnosis can effectively prevent these problems.

To provide the rationale for decision-making, \citep{melendez2014novel, melendez2015combining} attempted to combine instance-level responses for a bag classification, which could highlight abnormalities in the image.
However, these methods have poor instance-level accuracy and are inconsistent in MIL methods at the instance level \citep{kandemir2015computer, cheplygina2015label}.
One of the promising reasons for this is that MIL models may detect only the most abnormal parts of the image where multiple abnormalities are present \citep{cheplygina2019not}.
Recently, to alleviate the issue, \citep{pmlr-v80-ilse18a} introduced an attention-based MIL.
The attention mechanism was used to determine instances that trigger the bag label, called key instances \citep{liu2012key}.
Compared to the attention-based MIL, discriminative brain-region localization in AD studies can be considered a type of key instance detection to localize brain regions where the bag label is triggered.

We propose the BrainBagNet with a position-based gate (PG-BrainBagNet) for joint learning of pathological region localization and disease diagnosis.
As illustrated in Fig. \ref{fig:fig_1}, the proposed method starts with two branches, called the patch-level prediction branch and position-based gating branch. 
Given an MRI scan, the patch-level prediction branch extracts local features from a specific receptive field size (\ie, 3D patches) and produces patch-level responses.
As all MRI scans are aligned to a 3D template in image processing, the position information in 3D space can indicate brain regions.
The position information of the brain image is represented via coordinates in 3D Cartesian space.
Then, given the position information, the position-based gating branch identifies the brain regions where the class-discriminative responses consistently appear across subjects.
Whereas the existing approaches select patches before learning feature representation, the proposed position-based gating branch learns pathological region localization by linearly interacting with patch-level responses.
By leveraging the soft region proposals generated by the position-based gating branch, patch-level responses obtained in the patch-level prediction branch are converted to patch-level class evidence, which is transparently aggregated to determine an image-level response.
Transparent aggregation for the image-level response alleviates the difficulty of interpretation due to global and nonlinear interactions.
We evaluated the effectiveness of the proposed method on the Alzheimer's Disease Neuroimaging Initiative (ADNI) dataset. 
In addition, we conducted extensive ablation studies in both AD vs. NC and pMCI vs. sMCI binary classification tasks, called AD diagnosis and the MCI conversion prediction task, respectively.
The proposed method outperformed the comparison methods in both AD diagnosis and MCI conversion prediction tasks.
We identified the discriminative brain regions and patch-level class evidence in a weakly supervised learning manner and analyzed the changes according to patch size.

    
\section{Related works}
\label{sec:related_work}

\subsection{CNN-based Alzheimer's Disease Diagnosis}
The development of deep learning methods, including the CNN, has efficiently addressed multistep pipelines for handcrafted feature generation/extraction and logistic regression by training a model in an end-to-end manner \citep{korolev2017residual}.
Thus, studies on the accurate AD diagnosis based on the 3D CNN are underway by taking 3D whole-brain images as input \citep{basaia2019automated, fung2019alzheimer, pmlr-v116-liu20a}.
In particular, various architectures have been proposed for accurate AD diagnosis \citep{korolev2017residual, wang2020ensemble, jin2019attention, jin2020generalizable}.
For instance, \citep{pmlr-v116-liu20a} demonstrated the changes in disease identification performance according to various factors, such as the normalization layer, kernel size, network architecture width, and patient age.
For network architecture, the model introduced in \citep{pmlr-v116-liu20a} was designed to learn subtle changes in the brain by limiting the size reduction of feature maps during low-level feature extraction steps.

Moreover, the attention mechanism gradually became popular and widely employed in the CNN-based image recognition model to better explain network behavior and generate more discriminative feature representations \citep{zhang2019self, jetley2018learn, schlemper2019attention}.
In AD analysis, \citep{jin2019attention, jin2020generalizable} introduced an attention-based 3D convolutional network for disease identification and biomarker exploration.
The network architecture was designed based on the ResNet \citep{he2016deep}, and the attention module was embedded in the middle of the network.
By taking the extracted local features as input, the attention module produced spatial weights.
As described in the literature, the goal of the attention module is to represent the regional importance during end-to-end training.
Moreover, in backward propagation, the produced attention could work as a gradient filter.

As the early AD stages could only be identified using subtle local pathological cues, patch-level feature representations have been investigated to capture subtle local pathological changes more efficiently.
However, only a few brain regions could contain cues for disease identification; thus, an additional discriminative patch extraction procedure was required in advance.
Employing prior anatomical knowledge and the discriminative probability yielded by statistical approaches was the typical initial step for extracting discriminative patches \citep{tong2013multiple}.
Patch extraction has been improved through resampling schemes using Elastic Net \citep{janouvsova2012biomarker, tong2014multiple}.
Recently, a landmark discovery algorithm for AD diagnosis was introduced for discriminative patch extraction \citep{liu2018landmark}.
The algorithm started with a multivariate statistical test on training images performed using nonlinear registration.
A $p$-value map was obtained from the template space.
Based on the $p$-value map, landmarks were determined based on the size and number of patches and the minimum distance between patches.
Patches extracted from the landmark were used for diagnosis model training.
Moreover, as many 3D CNNs as the number of determined landmarks were configured to extract features for each patch.
The following fully connected layers employed concatenated patch-level representations for bag-level prediction.

However, discriminative region localization and patch extraction were performed independent of the diagnostic model, which could result in suboptimal diagnostic performance \citep{lian2018hierarchical}.
To alleviate this limitation, \citep{lian2018hierarchical} proposed a hybrid loss and pruning strategy based on the H-FCN. 
The proposed model extracted multiscale feature representations (\ie patch-, region-, and subject-level features) by employing the CNN. 
This hierarchical construction of the network architecture allows the trained model to identify the most informative patches and regions through hybrid loss.
Pruning less discriminative areas identified by hybrid loss could improve diagnostic results.
However, hybrid loss was defined by considering all patch- and region-level features belonging to the patient's MRI images as positive samples, although not all patches and regions would necessarily be affected by AD.
In addition, this approach still relies on predetermined landmarks for better classification performance in the initial stage.

Recently, a HybNet \citep{lian2020attention} was proposed to improve the H-FCN, which considers both global and local structural information.
Specifically, two branches were constructed: the global branch (GB) and local branch (LB). 
The subject-specific and intersubject-consistent discriminative region localization approaches were applied in the  GB and LB, respectively.
Both localizations were extracted by the pretrained fully convolutional network (FCN) backbone. 
The FCN backbone was trained for weakly supervised object localization (WSOL) and could generate a class activation map \citep{zhou2016learning}. 
Disease attention maps (DAMs) have been produced to represent subject-specific AD-related brain regions based on class activation map outcomes.
Moreover, the mean of the DAMs is calculated by simply averaging DAMs produced by the considerable samples in the training set.
Given the localization results in advance, the GB and LB were trained. 
The GB used DAMs, which represent subject-specific discriminative brain regions, as the spatial attention.
However, the LB was trained given patches extracted using intersubject-consistent discriminative brain regions represented by the mean DAM.
The patch extraction was performed identically to the H-FCN, but the mean DAM was used instead of the discriminative probability map obtained by the statistical test.
Moreover, this study indicated that the feature representations extracted from both branches could be complementary, and their fusion could improve classification performance.
Although additional information was used to address the shortcomings of H-FCN, the predetermination of patches may hamper the effectiveness of end-to-end learning of local feature extraction and diagnosis.

\subsection{AD-associated Brain-region Localization}
The AD-associated brain-region localization methods have been proposed and developed for various purposes.
Specifically, these methods boost the classification performance of diagnostic models, detect potential biomarkers in AD diagnosis, and better explain the behavior of deep learning networks.
These brain-region localization methods can be divided into two categories according to the information used in the localization: feature-based and position-based approaches.

First, feature-based approaches produce the brain-region localization result based on individual local features extracted from each brain image.
These can be further divided into supervised learning and weakly supervised learning approaches based on the learning strategy.
For supervised learning approaches, a relatively large patch or region of interest (ROI) extracted from the image was assigned the same annotation as the image-level annotation \citep{lee2019toward, qiu2020development}.
A model was trained to represent regional abnormalities by subject.
Regional outcomes were used as features to estimate the individual disease states.
Although the identified regional abnormalities can provide clinical evidence, the evidence was relatively coarse.
In addition, this supervised approach assumes that all regions in patients are affected by AD.
This fact has recently led to the application of weakly supervised learning and MIL.

Regarding weakly supervised learning approaches, \citep{zhou2016learning} proposed a representative WSOL method through an FCN with a global average pooling (GAP) layer and linear classifier.
Due to the linear property of the GAP and linear classifier, areas that contributed significantly to the predictions can be tracked.
This approach has evolved from several perspectives \citep{singh2017hide, yun2019cutmix}, and \citep{brendel2018bagnets} proposed bag-of-local-feature models, which provide patch-level class evidence by limiting the receptive field size of the topmost feature maps.
In the AD study, WSOL was employed in \citep{lian2018hierarchical} to localize AD-related structural abnormalities at a finer scale by training an additional 3D FCN.
Moreover, WSOL was applied to represent the regional importance for better feature representation \citep{li2019novel, lian2020attention}.
\citep{li2019novel} proposed an iterative learning framework leveraged by the localization result generated by WSOL.
Further, \citep{lian2020attention} introduced a subject-specific discriminative brain-region localization called a DAM.
For a similar purpose, an attention mechanism can be attached to a diagnostic model.
\citep{jin2019attention, jin2020generalizable} proposed an attention-based diagnosis model for joint learning of discriminative brain-region localization and disease identification.

Unlike feature-based region localization, position-based brain-region localization methods detect regions where significant differences appear between the AD and NC groups.
The identified brain regions are consistent across subjects so that this method could be called intersubject-consistent discriminative region localization \citep{lian2020attention}.
All sMRI scans are aligned to the same template in preprocessing; thus, all samples share the same 3D space.
This shared space allows a group comparison of local features, and the statistical test could generate a probability map representing the discriminative capacity.
In particular, this position-based localization method has been widely used in patch extraction for patch-level feature representation.
Data-driven pathological brain-region localization approaches have continued evolving as described in the previous section such as the statistical approach \citep{tong2014multiple, suk2014hierarchical}, landmark discovery \citep{tong2014multiple}, pruning strategy \citep{lian2018hierarchical}, and mean DAM \citep{lian2020attention}.
However, the existing patch extraction methods are performed independently of image-level diagnostic model outcomes.

Inspired by the recent patch-level analysis in AD diagnosis, we propose a framework that jointly learns pathological region localization and disease identification in an end-to-end manner.
In addition, final decision-making is conducted through the transparent aggregation of the patch-level responses, providing patch-level class evidence for decision-making.
To the best of our knowledge, this framework is the first for joint learning of position-based discriminative brain-region localization and disease identification in an end-to-end manner.

\section{Materials and Methods}
\begin{table}[tb]
\centering
\caption{Alzheimer's Disease Neuroimaging Initiative (ADNI) cohort and its corresponding demographic information.}
\label{tb:table_1}
\centering\scriptsize{
\begin{threeparttable}
\begin{tabular}{llllll}
    \toprule
	Datatset     & Category  & Gender (Male/Female)   & Age  & MMSE & Education  \\
	\toprule
	
	\multirow{4}{*}{ADNI1}
	&NC          &   119/112   & 76.0 $\pm$ 5.0      & 29.1 $\pm$ 1.0         & 16.1 $\pm$ 2.8      \\
	&sMCI        &   148/75   & 74.8 $\pm$ 7.7       & 27.3 $\pm$ 1.8         & 15.5 $\pm$ 3.2      \\      
	&pMCI        &   103/65   & 74.7 $\pm$ 7.0       & 26.6 $\pm$ 1.7         & 15.7 $\pm$ 2.8      \\
	&AD          &   103/97   & 75.6 $\pm$ 7.7       & 23.3 $\pm$ 2.0         & 14.7 $\pm$ 3.2      \\
	\midrule
	
	\multirow{4}{*}{ADNI2}
	&NC          &   97/107   & 73.4 $\pm$ 6.4       & 29.0 $\pm$ 1.2         & 16.6 $\pm$ 2.5                      \\
	&sMCI        &   151/123  & 71.3 $\pm$ 7.5       & 28.2 $\pm$ 1.6         & 16.3 $\pm$ 2.6                      \\      
	&pMCI        &   45/38   & 72.9 $\pm$ 7.2        & 27.1 $\pm$ 1,8         & 16.2 $\pm$ 2.3                       \\
	&AD          &   91/68   & 74.9 $\pm$ 8.1        & 23.1 $\pm$ 2.1         & 15.7 $\pm$ 2.7                       \\
	\bottomrule
		
\end{tabular}
\end{threeparttable}
}
\end{table}

\subsection{Subjects and Image Processing}
\label{sec:Dataset}
We used two public datasets (ADNI-1 and ADNI-2) from the {ADNI}\footnote{\url{http://adni.loni.usc.edu/}} cohort.
First, we collected the baseline brain sMRI scans and the diagnostic information from the datasets. 
Then, we removed the scans that appear in both ADNI-1 and ADNI-2 from ADNI-2, so that our dataset contains one sMRI scan for a subject.
The disease state of collected scans was categorized into three classes: NC, MCI, and AD.
We further divided each MCI subject into two classes for the MCI conversion prediction task.
If the patient corresponding to a baseline image had not been diagnosed with an AD class by 72 months, the image was labeled as the sMCI class.
In addition, images converted into the AD class within 36 months were labeled as the pMCI class.
In this process, MCI samples with reversion from the AD class to other classes were excluded from the dataset.
Overall, in ADNI-1, there were 231, 223, 168, and 200 sMRI scans for NC, sMCI, pMCI, and AD, respectively. 
For ADNI-2, there were 204, 274, 83, and 159 scans in the same order.
The demographics and clinical information are presented in Table \ref{tb:table_1}.

The prepared brain scans were processed using the following pipeline.
First, the brain extraction procedure was performed by the HD-BET brain extraction tool \citep{isensee2019automated} to remove areas other than brain images (\eg, neck, skull, and so on.).
Then, an affine registration was performed to linearly align each sMRI scan to the MNI152 template.
The process has been done through the FLIRT method in the {FSL}\footnote{\url{http://fsl.fmrib.ox.ac.uk/fsl/fslwiki}} package.
This process removed global linear differences over the images such as global translation, scale, and rotation differences. 
Furthermore, the process allowed all images to have identical spatial resolution ($1\times 1 \times 1 \text{mm}^3$).
Finally, the size of the processed brain scans was $193 \times 229 \times 193$, and each image was normalized through the mean and standard deviation of each image.


\begin{figure}[tb]
\centering
\includegraphics[width=1.0 \textwidth]{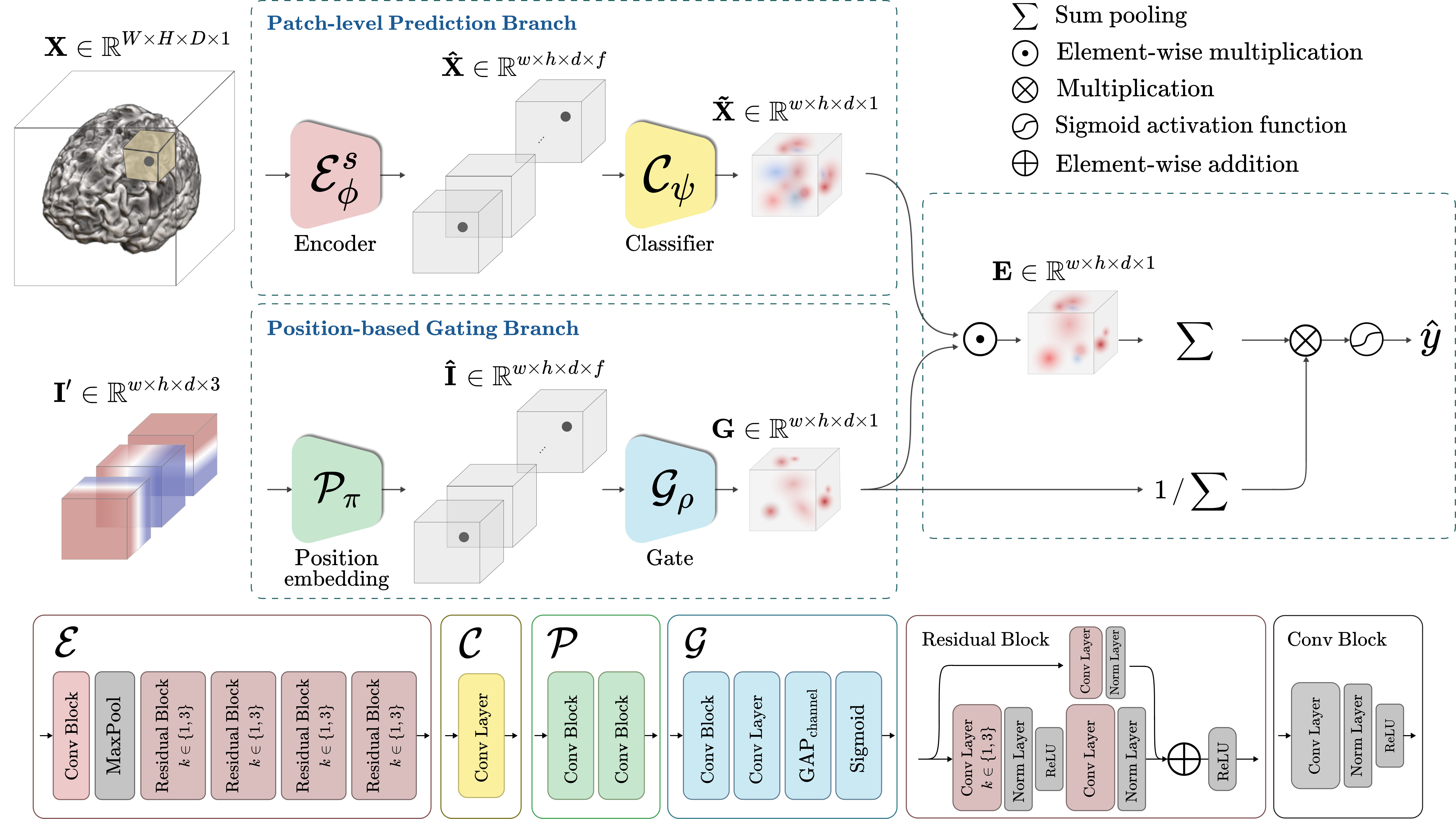}
\caption{Illustration of the proposed PG-BrainBagNet, including the encoder, classifier, position embedding, and gate network. Structural and position information is processed in two separate branches, given a magnetic resonance imaging scan and position indicator. The outcomes from the two branches are combined to represent patch-level class evidence and image-level disease probability.} 
\label{fig:fig_1}
\end{figure}

\begin{figure}[tb]
\centering
{
\begin{subfigure}[b]{0.25\textwidth}
\centering
\includegraphics[width=\linewidth]{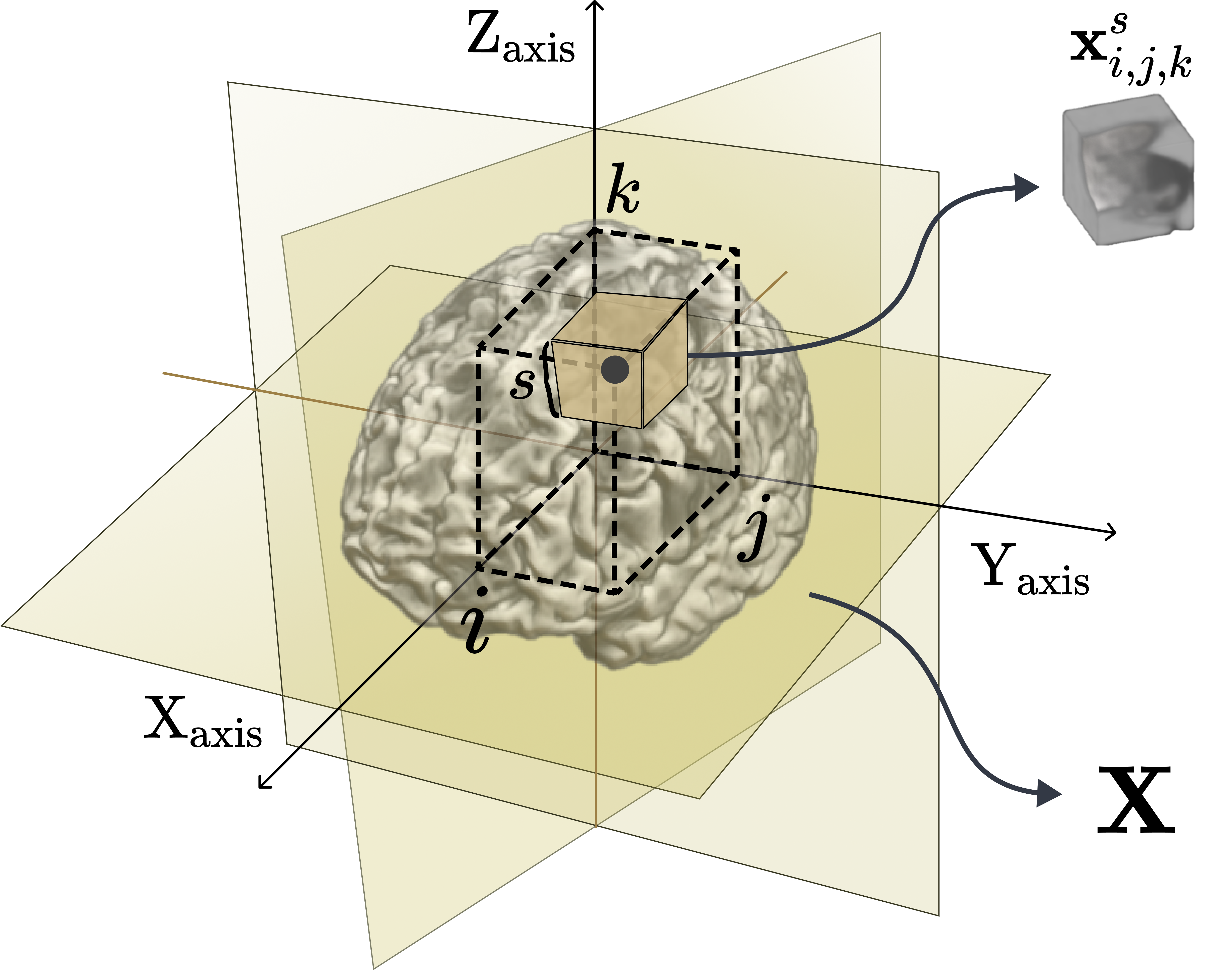}
\caption{Three-dimensional (3D) patch $\mathbf{x}^{s}_{i,j,k}$ comprising a 3D structural magnetic resonance imaging (sMRI) scan $\mathbf{X}$ on the Cartesian coordinate system.} 
\label{fig:fig_2_1}
\end{subfigure}
\hspace{1.cm}
\begin{subfigure}[b]{0.6\textwidth}
\centering
\includegraphics[width=\linewidth]{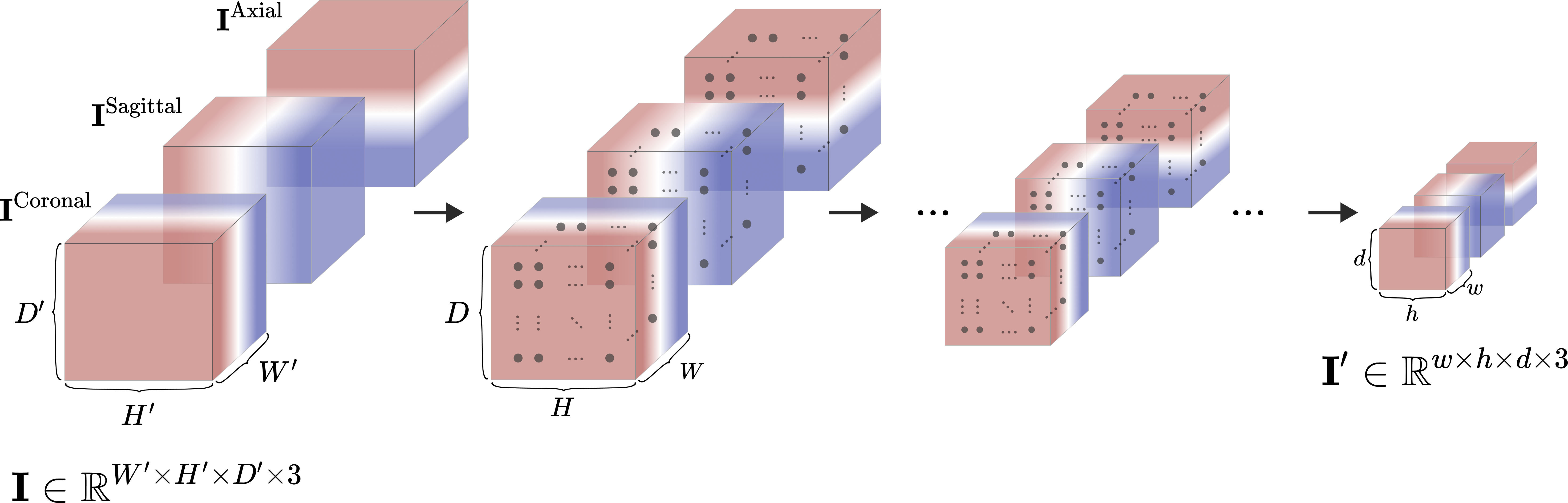}
\caption{Representation of coordinates in three-dimensional Cartesian space $\mathbf{I}$ and the extraction of position indicator $\mathbf{I}'$.}
\label{fig:fig_2_2}
\end{subfigure}
\caption{Illustration of (a) three-dimensional (3D) patch extracted in a specific position and size and (b) the position indicator extracted from the coordinates in 3D Cartesian space.}
\label{fig:fig_2}
}
\end{figure}

\subsection{Overview of Methods}

An overview of the proposed PG-BrainBagNet is presented in Fig. \ref{fig:fig_1}. 
There are four parameterized networks: the encoder, classifier, position embedding, and gate network.
These networks are organized in two branches, and each branch takes different inputs: $\mathbf{X}$ and $\mathbf{I}'$.
The input MRI scan $\mathbf{X}$ can be considered a set containing $M$ patches $\mathbf{X} = \{\mathbf{x}_1,\cdots,\mathbf{x}_M\}$. 
Moreover, $\mathbf{I}'$ is a representation for the patch position information: the position indicator.

The patch-level prediction branch comprises encoder and classifier networks.
First, we construct an encoder network that can adjust the patch size for feature extraction according to the receptive field size of the top-level feature maps.
The constructed encoder takes the whole brain image $\mathbf{X} \in \mathbb{R}^{W\times H\times D \times 1}$ as input and extracts the local features from 3D patches of size $s\times s\times s$.
Given the patch-level features $\hat{\mathbf{X}}$, the classifier network produces patch-level responses $\tilde{\mathbf{X}} \in \mathbb{R}^{w\times h\times d\times 1}$.
Thus, we obtained responses from $w\times h\times d$ number of patches distributed throughout the whole brain.
As the patch-level prediction branch shared the learning parameters over the spatial dimension, high-resolution input can be efficiently processed, and the overfitting problem can be avoided.
However, as the responses of patches are produced without considering their respective positions, there might be a crucial loss of information. 
The position-based gating branch can address this issue.

The MRI scans were processed by linear affine registration to a template; thus, all images share a 3D Cartesian space.
Through the 3D Cartesian coordinate system, a patch located in $(i,j,k)$ can be defined as $\mathbf{x}^{s}_{i,j,k}$, where $(i,j,k)$ denotes the coordinates in 3D space, as described in Fig. \ref{fig:fig_2_1}.
To reflect the position information in patch-level representation, we represent coordinates in 3D Cartesian space as a 4D tensor inspired by prior work \citep{liu2018intriguing} and introduce a scheme for extracting position information of patch-level responses, as presented in Fig. \ref{fig:fig_2_2}.
First, coordinates in 3D Cartesian space are represented as a 4D tensor $\mathbf{I} \in \mathbb{R}^{W'\times H'\times D'\times 3}$, which consists of three channels (\ie, coronal, sagittal, and axial axis).
Data augmentation, such as image translation and random cropping, can be applied in model training; thus, the spatial sizes of the 4D tensor and the input MRI scans may not be identical.
The representation of the 3D space coordinates indicates the brain-region position and allows extracting position information from which patch-level responses are obtained.
The extracted patch position information is denoted as a position indicator $\mathbf{I}'$.
Then, given the position indicator, the position-based gating branch consisting of the position embedding and gate network produces position-based responses.
The high responses represent the AD-related brain region.

Lastly, element-wise multiplication of the outcomes obtained from the two branches generates patch-level class evidence.
The image-level disease state is inferred by the transparent aggregation of the patch-level class evidence.
This process leads the model to be trained to detect patch-level class evidence by jointly learning AD-related local morphological changes and the regions where the discriminative changes sustainably appear.
Moreover, the patch-level class evidence indicates which patches contributed significantly to the model prediction.
In the following section, we present the details of these steps.

\subsection{Patch-level Features Extraction and Classification}
\label{sec:Encoder}
To handle local and morphological changes distributed in the whole brain, we constructed an encoder $\mathcal{E}^s_{\phi}$ and classifier network $\mathcal{C}_{\psi}$ parameterized with $\phi$ and $\psi$, respectively.
These networks comprise convolutional layers, which share learning parameters across the spatial dimension.
Thus, we effectively extract patch-level feature representation and patch-level responses with whole brain images as input by controlling the receptive field size of the topmost feature maps.
In particular, the kernel and stride sizes in the convolutional operator allow adjusting the patch sizes and distance between them.
Additionally, we employed this method to construct network architectures introduced in BagNets \citep{brendel2018bagnets}.
Based on BagNets, we reconstructed shallower encoders employing 3D convolutional layers.
The goal of configuring an encoder network $\mathcal{E}^s_{\phi}$ is to represent local features extracted from patches of size $s \times s\times s$.
In addition, the encoder network is configured so that the number of extracted patches and center position is the same regardless of patch size to compare the changes by size fairly.

Specifically, encoder network $\mathcal{E}^s_{\phi}$ consists of convolutional block, max pooling, and four residual blocks, as illustrated in Fig. \ref{fig:fig_1}.
All convolutional blocks in this study include sequential operators of the convolutional layer, instance normalization layer, and rectified linear unit (ReLU) activation function.
The kernel and stride size for the first convolutional block is set to $5\times 5\times 5$ and $2\times 2\times 2$, respectively.
For the following max-pooling layer, the kernel size is $3\times 3\times 3$, and the stride size is $2\times 2\times 2$.
The feature maps yielded by max pooling have a receptive field size of $9\times 9\times 9$.
Thus, if the receptive field size does not increase further, the encoder can extract features with a specific receptive field size of $9\times 9\times 9$ from the whole brain.
With nine as the minimum size, we constructed encoders based on five receptive field sizes according to the following residual blocks.
The residual blocks were modified based on the blocks used in ResNet18, the basic residual block \citep{korolev2017residual}.
Each residual block takes an argument $k\in \{1, 3\}$, representing a $k \times k\times k $ kernel size of the first convolutional layer, as described in Fig. \ref{fig:fig_1}.
The other convolutional layers in each residual block have point-wise convolution, enhancing the complexity in the representatives without increasing the receptive field size.
For the four sequential residual blocks that take [1, 1, 1, 1] as arguments, the receptive field size is not increased, and $\mathcal{E}^{9}_{\phi}$ denotes the encoder.
For $\mathcal{E}^{17}_{\phi}$, the sequential residual blocks take $k$ as [3, 1, 1, 1].
Likewise, $\mathcal{E}^{25}_{\phi}$, $\mathcal{E}^{41}_{\phi}$, and $\mathcal{E}^{57}_{\phi}$ take $k$ as [3, 3, 1, 1], [3, 3, 3, 1], and [3, 3, 3, 3], respectively.
For the normalization layer and nonlinear activation function in residual blocks, instance normalization layer and ReLU is employed, respectively.
Different kernel sizes in residual blocks can change the total number of local features and their center position of the receptive field, which might be additional variables to affect decision-making.
All convolutional layers in the encoder network involve replicating padding as much as the quotient resulting from dividing the kernel size by two for fair comparison by patch size.


Overall, we can achieve patch-level feature representation $\hat{\mathbf{X}}\in \mathbb{R}^{w \times h \times d \times f}$ for an individual MRI scan $\mathbf{X}$.
Patch-level features are represented in feature maps, where $w$, $h$, and $d$ denote the size of the feature map, and $f$ is the corresponding number of feature maps.
Therefore, $w\times h\times d$ number of patches are embedded in the $f$-dimensional vector space.
Then, the classifier network $\mathcal{C}_{\psi}$ converts the $f$-dimensional vector into a scalar for patch-level responses and produces  $\tilde{\mathbf{X}} \in \mathbb{R}^{w \times h \times d \times 1}$.
The patch-level responses $\tilde{\mathbf{X}} = (\tilde{x}_{1,1,1},\cdots,\tilde{x}_{i,j,k},\cdots,\tilde{x}_{w,h,d})$ comprises local responses $\tilde{x}_{i,j,k} \in \mathbb{R}$.
As both the encoder and classifier networks extract local responses in the specific receptive field size and share the extracting function over all spatial dimensions, each local response was extracted by a 3D patch in the brain without considering its position in the brain.

\subsection{Position-based Gate for AD-related Brain-region Localization}
\label{sec:Gate}

As all MRI scans were aligned in a 3D template in the image processing step, a 3D space can be shared and is applicable over the samples.
Moreover, all 3D patches are single-scale patches with 3D cubic shapes.
Thus, the patches distributed in the entire brain can be differentiated only by the patch position information, and the center patch position is the representative position information.
The simplest method to indicate positions is to use a one-hot representation.
However, this approach can be inefficient due to the numerous patches and ignores the volumetric position in 3D space.
This problem can be efficiently addressed using the Cartesian coordinate system.
Inspired by a representation proposed in \citep{liu2018intriguing}, we constructed a 3D complete translation invariance to specify a 3D Cartesian space.
The 3D Cartesian space coordinates could be represented in three channels, such as $\mathbf{I} \in \mathbb{R}^{W'\times H'\times D' \times 3}$.
Each coordinate channel is a Rank 1 tensor, such as $\mathbf{I}^{\text{Coronal}} \in \mathbb{R}^{W'\times H'\times D' \times 1}$, whose values are filled with 1's in the first coronal plane, 2's in its second coronal plane, 3's in the third coronal plane, and so on. 
The values from 1 to the number of coronal planes are normalized between -1 and 1.
Similarly, for the $\mathbf{I}^{\text{Sagittal}}$ and $\mathbf{I}^{\text{Axial}}$ coordinate channels, the values are filled through the sagittal and axial planes, respectively.
Therefore, these three coordinate channels are concatenated and result in a tensor $\mathbf{I} \in \mathbb{R}^{W'\times H'\times D'\times 3}$, which completely specifies the position of 3D Cartesian space.
Based on the position information in the space, we obtain the center position information for the patches used in the patch-level prediction branch.
The center position information is represented as a position indicator $\mathbf{I}'$.
The extraction of the position indicator is described in Fig. \ref{fig:fig_2_2}.
First, when using data augmentation such as image translation and cropping, the representation of 3D Cartesian space $\mathbf{I}$ should be transformed in the same way as the input transformation, which results in the same spatial dimension as the input MRI scan.
Then, based on the encoder network, the center positions of the receptive field are hierarchically extracted.
The extraction was performed by employing depth-wise convolution with nonparametric kernel weights, and all weights in the kernel were set to 0, but the center-position weight is 1.

In the end, the extracted position indicator $\mathbf{I}'\in \mathbb{R}^{w\times h\times d\times 3}$ is taken as input for the position-based gating branch.
The branch generates translation-dependent outcomes, which is not possible in the patch-level prediction branch.
As described in Fig. \ref{fig:fig_1}, parametric functions in position embedding $\mathcal{P}_{\pi}$ and gate network $\mathcal{G}_{\rho}$ consist of convolutional layers, and all convolutional layers in both networks are point-wise convolutions parameterized by $\pi$ and $\rho$.
In the position embedding network, the semantic feature representation $\hat{\mathbf{I}}$ were extracted to detect the task-oriented discriminative region by increasing the number of feature maps.
Furthermore, the number of output feature maps was decreased in the gate network to encode the semantic feature representation.
Finally, the remaining feature maps were averaged and activated by the sigmoid activation function to generate discriminative probability map $\mathbf{G} = (g_{1,1,1},\cdots,g_{i,j,k},\cdots,g_{w,h,d})$.
The discriminative probability map consists of $g_{i,j,k}\in [0, 1]$, representing the position-based response located in $(i,j,k)$. 
By constructing position indicator $\mathbf{I}'$ based on the representation of coordinates in the 3D Cartesian space, absolute positioning can be performed and shared over the MRI scans for each patch. 
Therefore, the trained position embedding and gate network represent the high response in the region where the AD-related morphological changes are consistently captured.

\subsection{Gate-based Pooling for Image-level Prediction}
\label{sec:pooling}
    
By considering a 3D whole brain to be a bag and considering the local features extracted from 3D patches distributed in the whole brain to be instances, the proposed framework can be considered an MIL framework.
In conventional MIL-based classification problems, nonparametric operators (\eg, max and mean) have been widely used to aggregate instance-level representation into bag-level representation. 
Just as \citep{brendel2018bagnets} introduced GAP for patch-level responses in aggregation, the mean operator has also been used as a representative aggregation function, especially when more than one instance is needed to identify a bag.
The mean operation can directly calculate the image-level response, $z$, as follows:
\begin{equation}
    z = \frac{1}{whd} \sum^{w,h,d}_{i,j,k=1} \Tilde{x}_{i,j,k}.
\end{equation}
A parametric operator constructed by a neural network was proposed in \citep{pmlr-v80-ilse18a} to detect key instances and aggregate the responses based on them.
Inspired by the aggregation method, we defined the image-level response by aggregating patch-level responses through position-based outcomes of gate network.
The element-wise multiplication between patch-level responses $\mathbf{\Tilde{X}}$ and discriminative probability map $\mathbf{G}$ results in patch-level class evidence $\mathbf{E}\in \mathbb{R}^{w\times h\times d\times 1}$.
The total amount of the discriminative brain region is unknown; thus, the normalization is performed based on the sum of the discriminative probability map so that the amount of the gated regions is independent of the diagnostic results.
The aggregation of patch-level class evidence infers image-level abnormality and is defined as follows:
\begin{equation}
    z = \frac{1}{\sum^{w,h,d}_{i,j,k=1} g_{i,j,k}} \sum^{w,h,d}_{i,j,k=1} e_{i,j,k}
    \text{,}
\end{equation}
where $e_{i,j,k} = g_{i,j,k}\Tilde{x}_{i,j,k}$.
The image-level response $z$ is directly activated by the posterior probability $\hat{y} = p(y|\mathbf{X})$ using the sigmoid activation function. 
The patch-level class evidence $\mathbf{E}$ directly reveals which patches made a significant contribution in the final decision, making the model transparent and interpretable.

\subsection{Joint Learning of Pathological Brain-region Localization and Disease Identification}
\label{sec:joint_training}
The overall parameters (\ie, $\phi$, $\psi$, $\pi$, and $\rho$) are trained based on the image-level classification objective.
To better train from the generalization perspective, the proposed models were trained using two additional techniques: label smoothing and balanced cross-entropy, referring to prior studies \citep{NEURIPS2019_f1748d6b, lin2017focal, he2019bag}.
The classification loss function is described as follows:
\begin{equation}
    \mathcal{L}_{\text{cls}} = -\beta y^{LS}\text{log}\hat{y}-(1-\beta)(1-y^{LS})\text{log}(1-\hat{y}),
\end{equation}     
where $y^{LS} \in \{0.1, 0.9\}$ is the modified target and $\beta$ is a hyperparameter addressing the imbalanced classification problem.
In addition, $\beta$ was set to the inverse class frequency.
Precisely, the function was calculated using the number of samples with negative annotation ($y=0$) divided by the total number of samples.
The gradient generated by classification loss updates the parameters, $\phi$, $\psi$, $\pi$, and $\rho$.

Moreover, the element-wise multiplication operation between $\mathbf{G}$ and $\mathbf{\Tilde{X}}$ allows both forward and backward propagation to be highly dependent on each other. 
However, in the early stages of training, randomly initialized parameters yielded both $\mathbf{\Tilde{X}}$ and $\mathbf{G}$.
To impose the framework to explore more discriminative brain region localization, we employ an entropy loss for maximization of entropy $\mathbf{G}$, as follows:
\begin{equation}
    H(\mathbf{G})  = - \frac{1}{whd}\sum^{w,h,d}_{i,j,k=1}(
    {g_{i,j,k}}\text{log}{g_{i,j,k}}
    +(1-g_{i,j,k})\text{log}(1-g_{i,j,k})
    ),
\end{equation}    
\begin{equation}
    \mathcal{L}_{\text{ent}}  = - H(\mathbf{G}).
\end{equation}
The gradient generated by the entropy loss is affected on parameters $\pi$ and $\rho$.
The final total loss function is defined using hyperparameter $\lambda$ to weigh the classification loss and entropy loss.
Our proposed network is trained in an end-to-end manner with the following loss function:
\begin{equation}
    \mathcal{L}_{\text{total}} = \mathcal{L}_{\text{cls}} + \lambda \mathcal{L}_{\text{ent}}.
\end{equation}


\section{Experimental Settings, Results, and Analysis}

\subsection{Comparative Deep Learning Methods}
Three sMRI-based deep learning approaches were adopted to compare the proposed method to state-of-the-art methods.
These approaches were sequentially called the 3D CNN, attention-based 3D ResNet, and HybNet and were reimplemented in PyTorch and trained on identical data using the proposed method.
The compared methods were summarized as follows:
\begin{itemize}
\item 3D CNN \citep{pmlr-v116-liu20a} : 
The CNN-based classifier was trained end-to-end to classify disease states without anatomical prior knowledge and a localization method.
We adopted a proposed model architecture for a fair comparison without clinical information, such as patients' age.
Given a randomly cropped 3D image as input, sequential convolutional blocks extracted local features, and the output feature maps were flattened. 
The flattening vector was employed as input to the classifier.
The size of cropped images for random cropping was set identically to the proposed method.
Based on this setting, we compared our method to the model trained with widening factor (WF) of 1 and 2.

\item Attention-based 3D ResNet \citep{jin2020generalizable} : 
This method had a similar goal: jointly learn AD-related brain-region detection and disease identification in an end-to-end manner.
However, two underlying differences exist.
One is that the AD-related brain regions were detected based on local features.
The other difference is the nonlinear interactions between weighted local features for image-level decision-making.
Specifically, the proposed model took the 3D whole brain as input.
The 3D residual blocks were stacked for local feature extraction.
The local features were pooled by using GAP, and the fully connected layer was followed for classification.
An additional attention module was attached in the middle of the feature extraction to detect the AD-related brain region.
This module generated spatial attention weights based on local features extracted in the middle of the network, and the attention was applied to local features.

\item Hybrid network (HybNet) \citep{lian2020attention} :
The HybNet consists of two branches constructed to capture 1) global structural information and 2) local structural information.
First, the FCN backbone was trained to generate the DAM and mean DAM to train a GB and LB, respectively.
The DAM was directly used as the attention in training the GB, whereas the mean DAM was used to determine the brain regions to extract patches.
The LB was trained based on the patches extracted from predetermined brain regions.
The following pruning and fine-tuning steps were performed as described in the literature.
Finally, two discriminative feature vectors obtained using the GB and LB were concatenated and used as input for the training fusion branch. 
The fusion branch comprised two subsequent fully connected layers followed by ReLU activation.

\end{itemize}

\subsection{Experimental Settings}
To validate the proposed models, we conducted five-fold cross-validation on the AD diagnosis (AD vs. NC) and MCI conversion prediction (pMCI vs. sMCI).
First, the preprocessed samples were randomly partitioned into five groups by disease states.
A single group was retained as the test samples and another group was used as validation samples to select an optimal model in the training process.
The remaining three groups were used as training samples.
The sizes of the training, validation, and testing datasets were 60\%, 20\%, and 20\% of the total size, respectively.
The cross-validation process was repeated five times, with each of the five groups used exactly once as the validation and testing data.
In this paper, two different tasks were studied: AD diagnosis and MCI conversion prediction.
For both the comparative and proposed models, we first trained models for AD diagnosis and transferred the trained parameters to initialize the network for the MCI conversion prediction task. 
This transfer learning process was performed considering that the two tasks are highly correlated, and the MCI conversion prediction task is more challenging than the other task.
In the AD diagnosis task, datasets of normal and AD patients were concatenated according to training, validation, and testing.
The dataset preparation for the MCI conversion prediction task was identical to the AD diagnosis.
Moreover, for transfer learning, the five partitions for AD diagnosis and MCI conversion prediction were paired.
Therefore, the parameter for the $k{_\text{th}}$ model for the MCI conversion prediction was initialized by the $k{_\text{th}}$ trained model for AD diagnosis.
The classification performances acquired in five-fold cross-validation were measured regarding the accuracy, sensitivity, specificity, and the area under the receiver operating characteristic (AUROC).

The proposed method required a predetermined 3D Cartesian space to indicate the brain regions where local features were extracted.
The 3D Cartesian space $\mathbf{I}$ was represented with the size of $193 \times 229\times 193 \times 3$.
To avoid the overfitting problem, we adopted a random cropping strategy for data augmentation with the size of $177 \times 213 \times 177$ in training; thus, the size of input $\mathbf{X}$ was $177 \times 213 \times 177$.
In evaluation, the images cropped in the center were used.
The number of output feature maps in the encoder network was set to 32, 32, 64, 128, and 256 for the first convolutional block and four residual blocks, respectively.
In the position-based gating branch, output feature maps for the positional embedding network were set at 128 and 256 and were reduced to 128 and 16 in the gate network.
In this study, $\lambda$ for the weight between classification and entropy losses was set to 0.01.
For model initialization, all weights for the AD diagnosis model were initialized using the He initialization method \citep{he2015delving} and optimized using the Adam optimizer \citep{adamOp}.
The total number of epoch was set to 200, and early stopping with 30 patience was used with four mini-batch sizes. 
For scheduling the learning rate, we adopted cosine annealing with the learning rate warm-up method, referred by \citep{he2019bag}. 
Specifically, the learning rate was linearly increased from 0 to 1e-4 within five epochs and was decreased as a cosine function to a learning rate of zero.

\begin{table}[tb]
\caption{Performance comparison on the Alzheimer's Disease Neuroimaging Initiative (ADNI) dataset in Alzheimer's disease (AD) diagnosis (AD vs. normal control (NC)).}
\label{tb:table_2}
\centering\scriptsize{
{
\begin{threeparttable}
\renewcommand*{\arraystretch}{0.8}
\begin{tabular}{lllll}
\toprule
Methods   & Accuracy & Sensitivity & Specificity & AUROC \\
\toprule
3D CNN (WF : 1) \citep{pmlr-v116-liu20a}         & 0.7992 $\pm$ 0.0368 & 0.7547 $\pm$ 0.0712 & 0.8360 $\pm$ 0.0347 & 0.8806 $\pm$ 0.0338\\
3D CNN (WF : 2)         & 0.8043 $\pm$ 0.0138 & 0.7800 $\pm$ 0.0465 & 0.8246 $\pm$ 0.0353 & 0.8861 $\pm$ 0.0058\\ 
\midrule
Attention-based 3D ResNet \citep{jin2020generalizable}  & 0.8270 $\pm$ 0.0262 & 0.7937 $\pm$ 0.0499 & 0.8547 $\pm$ 0.0482 & 0.9047 $\pm$ 0.0147\\ \midrule
HybNet (FCN) \citep{lian2020attention}         & {0.8346 $\pm$ 0.0162} & {0.7773 $\pm$ 0.0579} & 0.8822 $\pm$ 0.0223 & 0.8971 $\pm$ 0.0130\\
HybNet (GB)         & 0.8460 $\pm$ 0.0193 & 0.7828 $\pm$ 0.0330 & 0.8984 $\pm$ 0.0088 & 0.8974  $\pm$ 0.0165\\
HybNet (LB)         & 0.8258 $\pm$ 0.0320 & 0.7996  $\pm$ 0.0622 & 0.8477 $\pm$ 0.0458 & 0.9016 $\pm$ 0.0227\\
HybNet (fusion)          & 0.8322 $\pm$ 0.0460 & 0.8080  $\pm$ 0.0626 & 0.8522 $\pm$ 0.0343 & 0.9079 $\pm$ 0.0221\\
\midrule
BrainBagNet-9             & 0.7373 $\pm$ 0.0127 & 0.7213  $\pm$ 0.0544 & 0.7504 $\pm$ 0.0358 & 0.8092 $\pm$ 0.0292\\
FG-BrainBagNet-9      & 0.7474 $\pm$ 0.0112 & 0.7353  $\pm$ 0.0263 & 0.7575  $\pm$ 0.0104 & 0.8301  $\pm$ 0.0206\\
PG-BrainBagNet-9        & 0.8686 $\pm$ 0.0281 & 0.8412  $\pm$ 0.0359 & 0.8913  $\pm$ 0.0311 & 0.9376  $\pm$ 0.0250\\  
\midrule
BrainBagNet-17            & 0.7525 $\pm$ 0.0251 & 0.7465  $\pm$ 0.0344 & 0.7573  $\pm$ 0.0544 & 0.8264  $\pm$ 0.0218\\
FG-BrainBagNet-17      & 0.8282 $\pm$ 0.0238 & 0.8162  $\pm$ 0.0366 & 0.8381  $\pm$ 0.0388 & 0.8925  $\pm$ 0.0220\\  
PG-BrainBagNet-17        & 0.8775 $\pm$ 0.0272 & 0.8384  $\pm$ 0.0409 & \textbf{0.9099}  $\pm$ 0.0224 & 0.9397  $\pm$ 0.0174\\  \midrule
BrainBagNet-25            & 0.8232 $\pm$ 0.0172 & 0.7882  $\pm$ 0.0315 & 0.8521  $\pm$ 0.0190 & 0.8910  $\pm$ 0.0199\\
FG-BrainBagNet-25      & 0.8446 $\pm$ 0.0245 & 0.8355  $\pm$ 0.0431 & 0.8522  $\pm$ 0.0379 & 0.9088  $\pm$ 0.0186\\
PG-BrainBagNet-25        & 0.8724 $\pm$ 0.0253 & 0.8329  $\pm$ 0.0457 & 0.9053  $\pm$ 0.0224 & 0.9391  $\pm$ 0.0111\\   
\midrule
BrainBagNet-41            & 0.8648 $\pm$ 0.0434 & 0.8413  $\pm$ 0.0677 & 0.8844  $\pm$ 0.0347 & 0.9291  $\pm$ 0.0343\\
FG-BrainBagNet-41      & 0.8749 $\pm$ 0.0304 & \textbf{0.8523}  $\pm$ 0.0329 & 0.8936  $\pm$ 0.0352 & 0.9199  $\pm$ 0.0292\\
PG-BrainBagNet-41        & \textbf{0.8825} $\pm$ 0.0127 & \textbf{0.8523}  $\pm$ 0.0144 & 0.9076  $\pm$ 0.0131 & \textbf{0.9403}  $\pm$ 0.0120\\ 
\midrule
BrainBagNet-57            & 0.8596 $\pm$ 0.0526 & 0.8354  $\pm$ 0.0724  & 0.8797  $\pm$ 0.0384 & 0.9137  $\pm$ 0.0310\\
FG-BrainBagNet-57      & 0.8497 $\pm$ 0.0198  & 0.8383  $\pm$ 0.0386  & 0.8592  $\pm$ 0.0302  & 0.9268  $\pm$ 0.0229\\ 
PG-BrainBagNet-57        & 0.8623 $\pm$ 0.0261 & 0.8355  $\pm$ 0.0374 & 0.8845  $\pm$ 0.0286 & 0.9371  $\pm$ 0.0196\\
\bottomrule
\end{tabular}
\begin{tablenotes}
    \scriptsize
    \item (WF: widening factor, GB: global branch, LB: local branch, FG: feature-based gate, PG: position-based gate)
\end{tablenotes}
\end{threeparttable}
}}
\end{table}

\begin{table}[tb]
\caption{Performance comparison on the Alzheimer's Disease Neuroimaging Initiative (ADNI) dataset in mild cognitive impairment (MCI) conversion prediction (progressive MCI vs. stable MCI).}
\label{tb:table_3}
\centering\scriptsize{
{
\begin{threeparttable}
\renewcommand*{\arraystretch}{0.8}
\begin{tabular}{lllll}
\toprule
Methods   & Accuracy & Sensitivity & Specificity & AUROC \\
\toprule
3D CNN (WF : 1)\citep{pmlr-v116-liu20a}         & 0.6886 $\pm$ 0.0275 & 0.2789  $\pm$ 0.0980 & 0.8953  $\pm$ 0.0664 & 0.7016  $\pm$ 0.0146\\
3D CNN (WF : 2)         & 0.6752 $\pm$ 0.0188 & 0.2391  $\pm$ 0.1278 & 0.8952  $\pm$ 0.0849 & 0.6920  $\pm$ 0.0239\\
\midrule
Attention-based 3D ResNet \citep{jin2020generalizable}  & 0.7018 $\pm$ 0.0292 & 0.2668  $\pm$ 0.0476 & \textbf{0.9215}  $\pm$ 0.0431 & 0.7204  $\pm$ 0.0356\\  
\midrule
HybNet (FCN) \citep{lian2020attention}         & 0.6979 $\pm$ 0.0187 & 0.4308  $\pm$ 0.0699 & 0.8329  $\pm$ 0.0504 & 0.7254  $\pm$ 0.0242\\
HybNet (GB)         & 0.6925 $\pm$ 0.0210 & 0.5096  $\pm$ 0.0764 & 0.7848  $\pm$ 0.0293 & 0.7286  $\pm$ 0.0223\\ 
HybNet (LB)         & 0.6939 $\pm$ 0.0391 & 0.3509  $\pm$ 0.1197 & 0.8672  $\pm$ 0.0195 & 0.7106  $\pm$ 0.0501\\ 
HybNet (fusion)          & 0.7018 $\pm$ 0.0143 & 0.4026  $\pm$ 0.0525 & 0.8531  $\pm$ 0.0392 & 0.6758  $\pm$ 0.0275\\ 
\midrule
BrainBagNet-9              & 0.6430 $\pm$ 0.0305 & 0.6652  $\pm$ 0.0536 & 0.6319  $\pm$ 0.0392 & 0.7159  $\pm$ 0.0411\\
FG-BrainBagNet-9       & 0.6430 $\pm$ 0.0328 & 0.6613  $\pm$ 0.0620 & 0.6338  $\pm$ 0.0399 & 0.7159  $\pm$ 0.0474\\
PG-BrainBagNet-9         & \textbf{0.7151} $\pm$ 0.0231 & 0.7051  $\pm$ 0.0546 & 0.7203  $\pm$ 0.0433 & \textbf{0.7726}  $\pm$ 0.0092\\  
\midrule
BrainBagNet-17             & 0.6551 $\pm$ 0.0255 & 0.6932  $\pm$ 0.0807 & 0.6360  $\pm$ 0.0303 & 0.7162  $\pm$ 0.0236\\
FG-BrainBagNet-17       & 0.6791 $\pm$ 0.0128 & 0.7131  $\pm$ 0.0395 & 0.6620  $\pm$ 0.0091 & 0.7565  $\pm$ 0.0075\\
PG-BrainBagNet-17         & 0.6965 $\pm$ 0.0045 & 0.6616  $\pm$ 0.0572 & 0.7142  $\pm$ 0.0313 & 0.7629  $\pm$ 0.0103\\
\midrule
BrainBagNet-25             & 0.6750 $\pm$ 0.0459 & 0.6893  $\pm$ 0.0635 & 0.6678  $\pm$ 0.0929 & 0.7269  $\pm$ 0.0320\\
FG-BrainBagNet-25       & 0.6885 $\pm$ 0.0133 & 0.6613  $\pm$ 0.0633 & 0.7022  $\pm$ 0.0177 & 0.7483  $\pm$ 0.0132\\
PG-BrainBagNet-25         & 0.6912 $\pm$ 0.0283 & 0.6498  $\pm$ 0.0648 & 0.7123  $\pm$ 0.0593 & 0.7526  $\pm$ 0.0313\\
\midrule
BrainBagNet-41             & 0.6911 $\pm$ 0.0088 & 0.7093  $\pm$ 0.0635 & 0.6821  $\pm$ 0.0359 & 0.7510  $\pm$ 0.0261\\
FG-BrainBagNet-41       & 0.6872 $\pm$ 0.0179 & 0.7016  $\pm$ 0.0849 & 0.6801  $\pm$ 0.0264 & 0.7474  $\pm$ 0.0164\\
PG-BrainBagNet-41         & 0.6953 $\pm$ 0.0437 & 0.7055  $\pm$ 0.0605 & 0.6902  $\pm$ 0.0537 & 0.7549  $\pm$ 0.0479\\  
\midrule
BrainBagNet-57             & 0.6831 $\pm$ 0.0290 & \textbf{0.7176}  $\pm$ 0.0873 & 0.6659  $\pm$ 0.0492 & 0.7485  $\pm$ 0.0409\\
FG-BrainBagNet-57       & 0.7006 $\pm$ 0.0190 & 0.6462  $\pm$ 0.1138 & 0.7283  $\pm$ 0.0511 & 0.7639  $\pm$ 0.0170\\
PG-BrainBagNet-57         & 0.6805 $\pm$ 0.0211 & 0.6893  $\pm$ 0.0626 & 0.6761  $\pm$ 0.0248 & 0.7552  $\pm$ 0.0149\\
\bottomrule
\end{tabular}
\begin{tablenotes}
    \scriptsize
    \item (WF: widening factor, GB: global branch, LB: local branch, FG: feature-based gate, PG: position-based gate)
\end{tablenotes}
\end{threeparttable}
}}
\end{table}

\subsection{Comparison of the Classification Performance}
Tables \ref{tb:table_2} and \ref{tb:table_3} present the comparison of classification performance for the AD diagnosis and MCI conversion prediction task, respectively. 
Each table lists the classification results of the state-of-the-art methods and our method. 
BrainBagNet-$s$ processes a single patch size $s$, and the patch-level responses are aggregated using the GAP operation.
The FG-BrainBagNet indicates a BrainBagNet with a feature-based gate inspired by an attention-based MIL framework \citep{pmlr-v80-ilse18a}.
Instead of using position information, local features generated by the encoder network were used as input to the gate network.
Lastly, the PG-BrainBagNet is the proposed method.

In the AD diagnosis task, the best classification performance in the experimental setting appeared in PG-BrainBagNet-41.
Compared to state-of-the-art methods, outperforming classification accuracy was observed. 
The highest and lowest margins for the mean accuracy were 8.33\text{\%} (vs. 3D CNN (WF:2)) and 3.65\text{\%} (vs. HybNet (GB)), respectively.
In addition, the position-based gate method yields an increase in classification performance compared to BrainBagNets.
In particular, the position-based gate method increased the classification performance of BrainBagNet with a small patch size by a large margin, whereas feature-based gates did not.
Specifically, for PG-BrainBagNet-9, the accuracy increased by 13.13\text{\%} (vs. BrainBagNet-9).
Although classification performance of BrainBagNets increased as the patch size increased, PG-BrainBagNets did not exhibit significant differences according to patch size.
Therefore, the reason that BrainBagNets performed classification poorly when using a small patch size might be that the whole-brain image contains many patches unrelated to the brain disease.
In addition, the improvement in classification performance indicated that the AD-related regions were effectively gated by the proposed method.

In the MCI conversion prediction task, the best prediction result appeared in PG-BrainBagNet-9.
In the comparison of the results using the state-of-the-art methods, the maximum and minimum differences in the mean accuracy were 3.99\text{\%} (vs. 3D CNN (WF:2)) and 1.33\text{\%} (vs. attention-based 3D ResNet and HybNet (fusion)), respectively.
In terms of patch sizes, the classification performance of BrainBagNets increased as the patch size increased.
Additionally, feature-based gate methods could not improve the classification results.
However, the position-based gate method yielded improvements when a small patch size was used, especially when the patch size was 9, 17, or 25.
Furthermore, the classification results were consistently increased as the patch size was reduced.
As the receptive field size was limited, local feature representations were forced to extract the local brain changes rather than global structural changes.
The results implied the brain-region localization method based on the position provided highly informative results.
In addition, we observed the importance of capturing subtle changes for the early detection of AD.
The MCI conversion prediction performance obtained from the proposed models trained from scratch and those trained through transfer learning is compared in the supplementary A.

\begin{figure}[tb]
\centering
\includegraphics[width=0.8\textwidth]{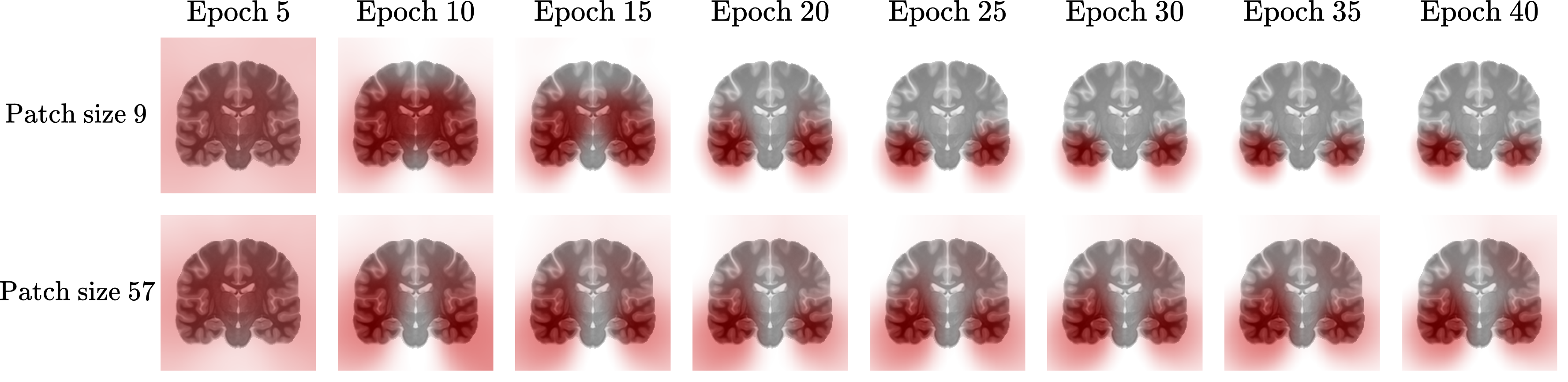}
\caption{Changes in the output of the proposed position-based gating branch according to the number of training epochs and patch size used in model training.}
\label{fig:fig_3}
\end{figure}

\begin{figure}[tb]
\centering
\includegraphics[width=0.8\textwidth]{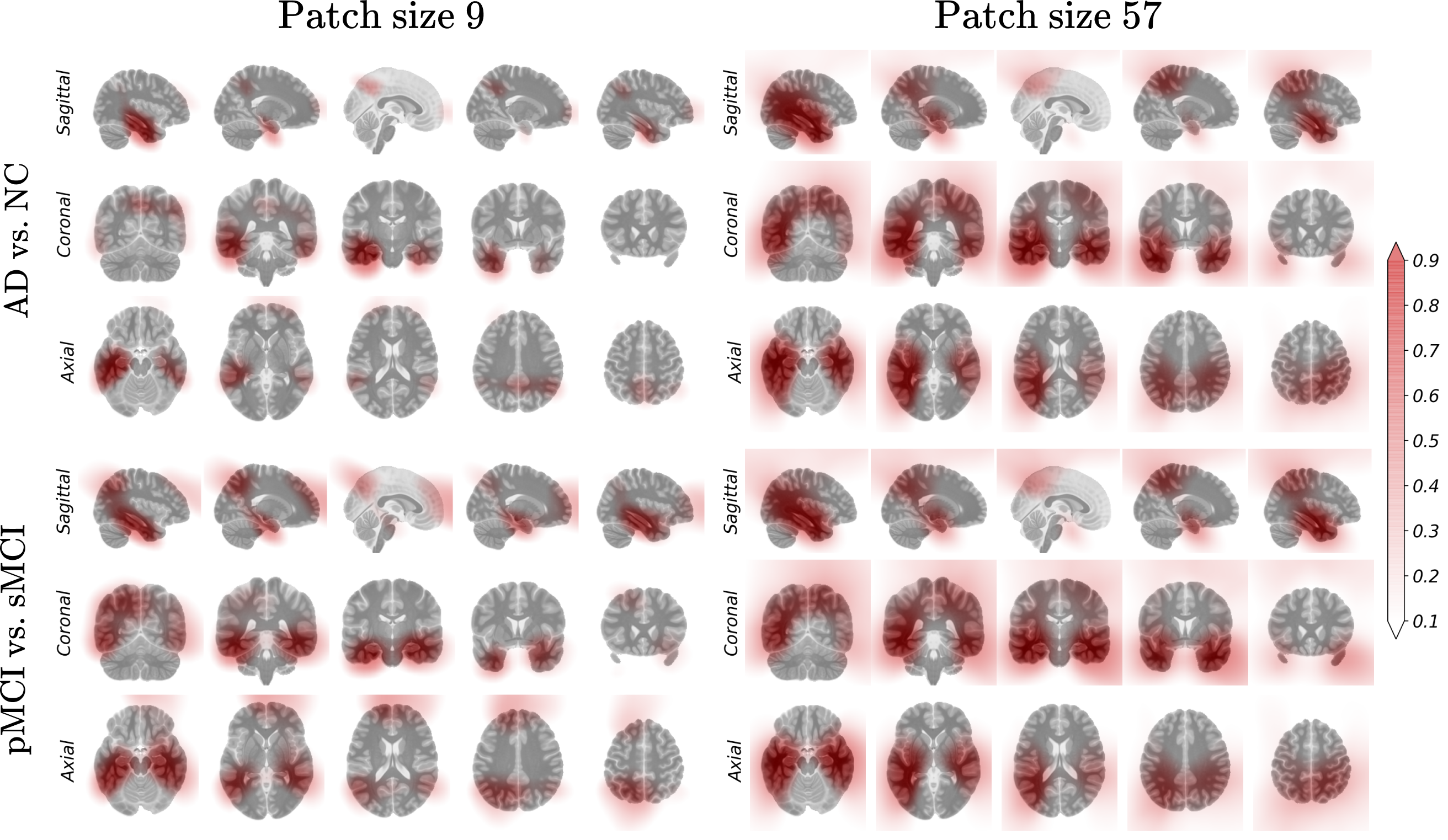}
\caption{Difference in the output of the trained position-based gating branch depending on the task and patch size used in model training.}
\label{fig:fig_4}
\end{figure}

\subsection{Result of Discriminative Brain-region Localization}
We analyzed discriminative probability maps $\mathbf{G}$ produced by the proposed position-based gating branch.
For better visualization, linear interpolation was performed and overlaid with the MNI template.
The changes in the discriminative probability map by the learning epoch are described in Fig. \ref{fig:fig_3}, and the results for the more varied patch sizes are presented in the supplementary B.
In the learning process with a small patch size, localization helped the diagnostic model extract finer features; thus, the gating branch could provide better localization results based on the features extracted by the diagnostic model.
In model training using a large patch size, we observed that the changes in the output of the gate network converged in an earlier epoch than that using a small patch.
The differences in the output of the gating branch from the trained model according to the patch sizes and tasks are compared in Fig. \ref{fig:fig_4}.
First, high responses were distributed in anatomically meaningful areas such as the hippocampal, temporal, and parietal lobe areas.
When the model was limited to increase the receptive field size, the gate network represented the weight in the sparse regions only.
As the patch size increased, regions with high responses were captured in the overall images.
In this context, the proposed method employing small patches is sensitive to localization results and requires proper localization.
While there were no significant changes by tasks, we observed that the highlighted regions were dispersed especially in the model trained using small patches for MCI conversion prediction.
From the perspective of discriminative brain regions, this result implies a high correlation between AD diagnosis and MCI conversion prediction task, but the difficulty of the MCI conversion prediction task reduced the model certainty in discriminative brain region localization.

\begin{figure}[tb]
\centering
\begin{subfigure}[b]{0.6\textwidth}
\centering
    \includegraphics[width=\linewidth]{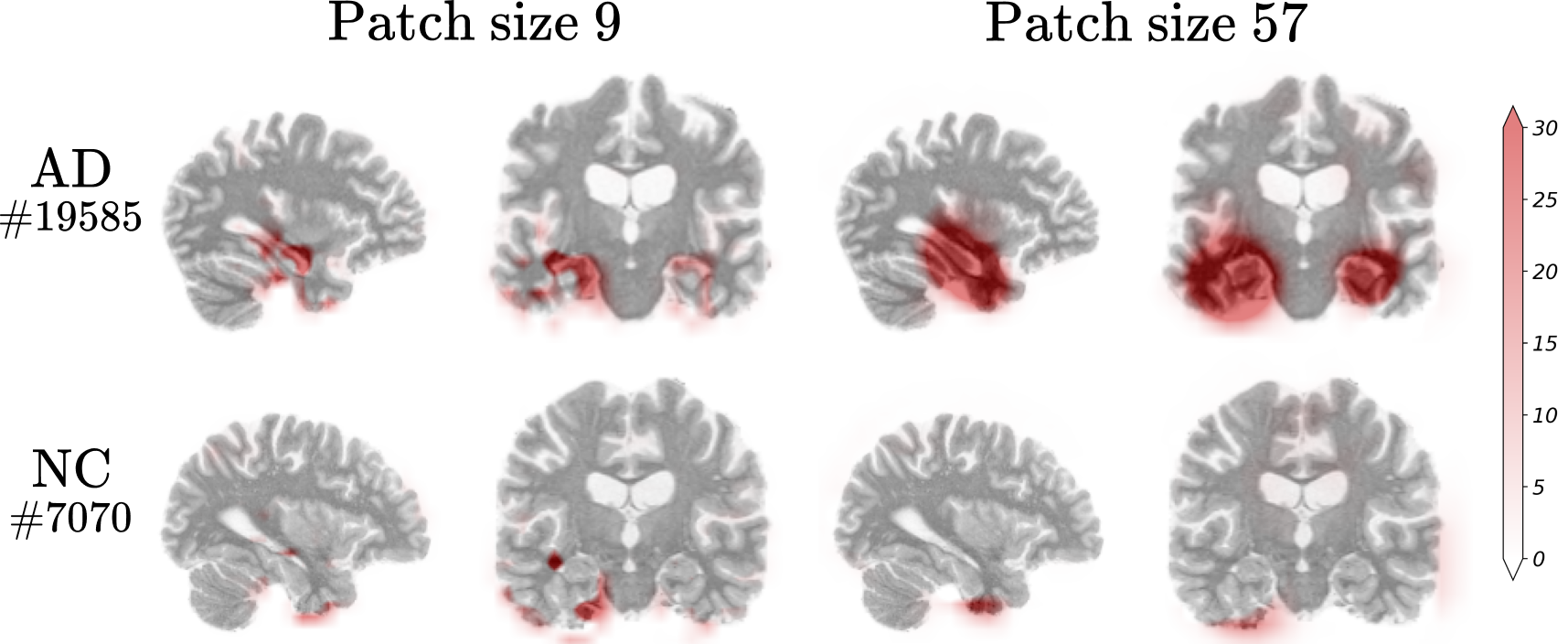}
\caption{AD vs. NC.}
\label{fig:fig_5_1}
\end{subfigure}
\par\bigskip
\begin{subfigure}[b]{0.6\textwidth}
\centering
    \includegraphics[width=\linewidth]{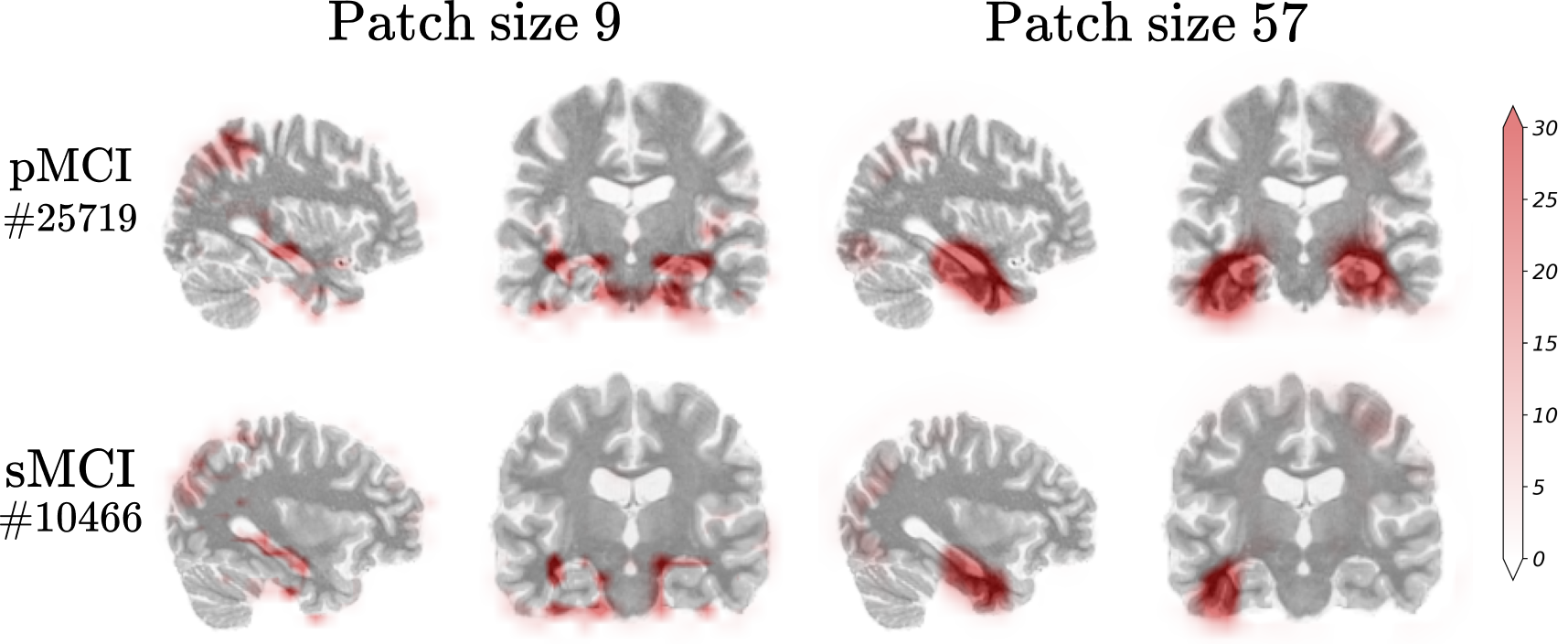}
\caption{pMCI vs. sMCI.}
\label{fig:fig_5_2}
\end{subfigure}
\caption{Difference in the positive patch-level class evidence produced by the proposed method according to disease state and patch size used in model training. Each row indicates one sample per disease state, where the number next to the \# denotes the corresponding image ID of an input MRI scan.}

\label{fig:fig_5}
\end{figure}

\subsection{Identified Patch-level Class Evidence}
The proposed method performed image-level decision-making by transparently aggregating patch-level class evidence.
Thus, higher patch-level class evidence results in a greater contribution to prediction.
As patch-level class evidence is created from individual 3D MRI scans, it provides an individualized analysis for AD progression.
Positive values were extracted, and linear interpolation was applied to analyze the local evidence for positive class, such as AD and pMCI.
The obtained 3D image was overlaid with the input image.
The results for one sample per disease state are described in Fig. \ref{fig:fig_5}.
First, we observed much more positive class evidence by the disease progression.
We also observed that the overall patch-level class evidence was consistently captured in the hippocampal region, temporal lobe, and parietal lobe.
However, even with the same 3D image as input, apparent differences depend on the patch size used in model training.
A trained model could detect local class evidence in the subtle brain atrophy around the hippocampal and parietal lobe area when the patch size was small.
The model trained using large patches seemed to detect cues from relatively coarse structural changes compared with the model trained using small patches. 
Results for a larger number of coronal, sagittal, and axial planes are presented in Supplementary C.
In addition, the visualization of patch-level class evidence for the more varied patch sizes and additional samples is depicted in supplementary D.

\section{Discussion}
\label{sec:discussion}

\subsection{Quantitative Analysis of Discriminative Regions by Patch Size}
\begin{figure}[tb]
\centering
\begin{subfigure}[b]{0.4\textwidth}
\centering
\includegraphics[width=\textwidth]{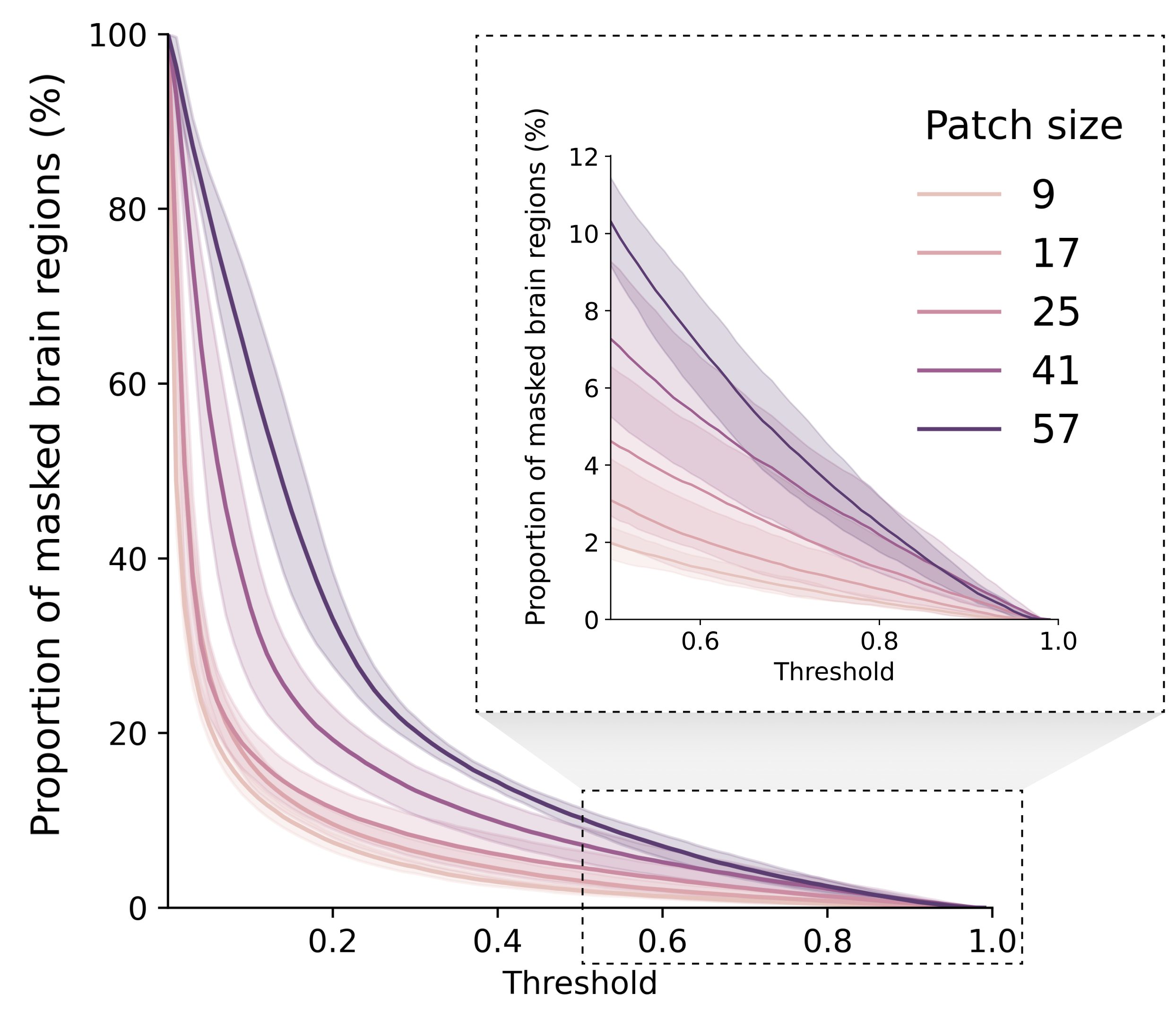}
\caption{AD vs. NC.} 
\label{fig:fig_6_1}
\end{subfigure}
\hspace{0.5cm}
\begin{subfigure}[b]{0.4\textwidth}
\centering
\includegraphics[width=\textwidth]{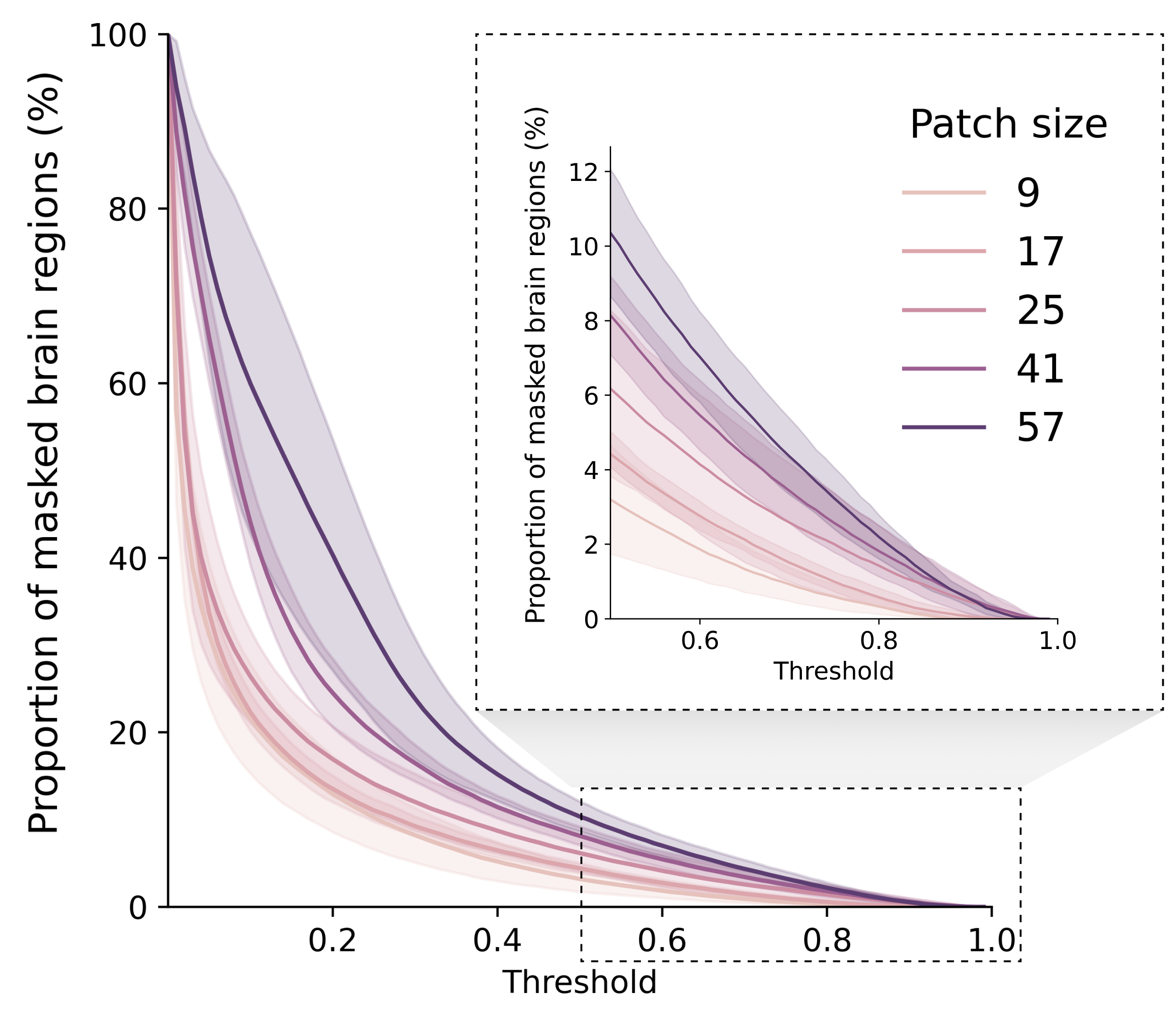}
\caption{pMCI vs. sMCI.}
\label{fig:fig_6_2}
\end{subfigure}
\caption{Difference in the proportion of masked brain regions according to the task, threshold, and patch size used in model training. The region where the output of the proposed gate network is higher than a given threshold was masked.}
\label{fig:fig_6}
\end{figure}

The existing AD analysis using patch-level feature representation inevitably extracted discriminative patches before performing feature extraction for diagnostic model training.
Moreover, patch extraction requires various hyperparameter adjustments, such as the patch size, number of patches to extract, and so on.
While the overall hyperparameters were highly correlated, this has been ignored in the patch extraction process.
For instance, as the patch size increases, the number of patches containing discriminative features increases.
Unlike previously proposed patch extraction \citep{tong2014multiple, suk2014hierarchical, lian2020attention}, the proposed method learned and probabilistically indicated the importance of overlapped and widely distributed patches in the whole brain.
We can perform a quantitative analysis of masked brain regions through a threshold between 0 and 1.
The region where values were higher than a given threshold was masked, and the proportion of masked regions was calculated.
As described in Fig. \ref{fig:fig_6}, by increasing the threshold from 0 to 1, the proportion of masked regions was sharply decreased for both prediction tasks of AD vs. NC and pMCI vs. sMCI.
When the thresholds were either small or large, the number of remaining patches varied according to patch size.
For example, the masked region difference between the AD diagnosis model with a patch size of 9 and the model with a patch size of 57 was about five times the 0.5 threshold.
This results revealed the relative sparsity of the discriminative brain regions in dealing with small patches in both AD diagnosis and MCI conversion prediction.
Furthermore, this sparsity explained the reason for extracting small discriminative patches, which was challenging and vital.

\begin{table}[!t]
\caption{Comparison of the classification accuracy through Monte Carlo dropout using four types of drop probability.}
\label{tb:table_4}
\centering\scriptsize{
{\begin{tabular}{cl|cccc}
\toprule
\multicolumn{2}{c}{Drop probability} \vline &$p_{i,j,k}=0.0$ &$p_{i,j,k}=0.5$ &$p_{i,j,k}=(g_{i,j,k})$ & $p_{i,j,k}=(1-g_{i,j,k})$  \\ 
\midrule
\multirow{5}{*}{AD vs. NC}
& Patch 9  &     0.8686$\pm$0.0281&     0.8699$\pm$0.0293&     0.5708$\pm$0.0341&      0.8132$\pm$0.0196 \\
& Patch 17 &     0.8775$\pm$0.0272&     0.8775$\pm$0.0307&     0.6252$\pm$0.0942&      0.8334$\pm$0.0252 \\
& Patch 25 &     0.8724$\pm$0.0253&     0.8686$\pm$0.0244&     0.6495$\pm$0.1222&      0.8737$\pm$0.0297 \\
& Patch 41 &     0.8825$\pm$0.0127&     0.8838$\pm$0.0157&     0.5318$\pm$0.0745&      0.8712$\pm$0.0059 \\
& Patch 57 &     0.8623$\pm$0.0261&     0.8648$\pm$0.0283&     0.6112$\pm$0.0644&      0.8711$\pm$0.0226 \\
\midrule
\multirow{5}{*}{pMCI vs. sMCI}
& Patch 9  &     0.7151$\pm$0.0231&     0.7098$\pm$0.0276&     0.5699$\pm$0.0759&      0.7220$\pm$0.0288 \\    
& Patch 17 &     0.6965$\pm$0.0045&     0.7005$\pm$0.0122&     0.5498$\pm$0.1178&      0.6979$\pm$0.0174 \\    
& Patch 25 &     0.6912$\pm$0.0283&     0.6885$\pm$0.0271&     0.4734$\pm$0.0791&      0.7153$\pm$0.0261 \\    
& Patch 41 &     0.6953$\pm$0.0437&     0.6940$\pm$0.0431&     0.4721$\pm$0.1174&      0.7259$\pm$0.0241 \\    
& Patch 57 &     0.6805$\pm$0.0211&     0.6819$\pm$0.0207&     0.4961$\pm$0.1162&      0.6737$\pm$0.0365 \\   
\bottomrule
\end{tabular}}}
\end{table}

\subsection{Stochastic Representation of Discriminative Regions}
To analyze the stochastic representation of discriminative brain regions, we evaluated the classification performance of the trained PG-BrainBagNet by employing the Monte Carlo dropout \citep{pmlr-v48-gal16}.
By randomly dropping $g_{i,j,k}$ with a probability of $p_{i,j,k} \in \{0.0,\text{ } 0.5,\text{ } g_{i,j,k},\text{ } (1-g_{i,j,k})\}$, we generated four types of randomly dropped $\mathbf{G}' \in \mathbb{R}^{w\times h\times d\times 1}$.
Based on $\mathbf{G}'$, we inferred $\hat{y}'$ for image-level prediction.
The final decision was made through soft voting results obtained through 100 trials.
The classification accuracy observed in the five-fold cross-validation is listed in Table \ref{tb:table_4}.

First, when the drop probability is 0, it is identical to the results without Monte Carlo dropout, which are presented in Tables \ref{tb:table_2} and \ref{tb:table_3}.
If the drop probability increased to 0.5, the prediction was made using the randomly dropped patch-level class evidence.
The best classification results are exhibited by patch sizes 41 and 9 in the AD diagnosis and MCI conversion prediction task, respectively.
Although the decision was made using randomly dropped local class evidence, the soft voting results show equivalent the classification accuracy with 0 drop probability.
When the drop probability was set as $g_{i,j,k}$, the model loses local class evidence appearing in the discriminative brain regions with high probability.
Therefore, regardless of the patch size, the classification result was similar to the chance ratio of a binary classification.
Lastly, when the drop probability was $1-g_{i,j,k}$, the model lost local class evidence appearing in the discriminative brain regions with low probability. 
However, the evidence from regions with relatively low $g_{i,j,k}$, such as the parietal and frontal lobes, might be dropped with high probability.
In the AD diagnosis task, we observed that the information generated in the regions decreased the classification accuracy when small patches were used. 
We speculated that the subtle changes captured in the parietal and frontal lobes could provide informative cues for AD diagnosis.
We observed increased classification accuracy in the MCI conversion prediction task when patch sizes of 9, 25, and 41 were used.
This result implies that the discriminative region representation $\mathbf{G}$ might contain uncertainty across subjects and that estimating the posterior probability of $y$ using Bayesian approaches could lead to better performance in predicting MCI conversion.

\begin{figure}[tb]
\centering
\begin{subfigure}[b]{0.8\textwidth}
\centering
\includegraphics[width=\textwidth]{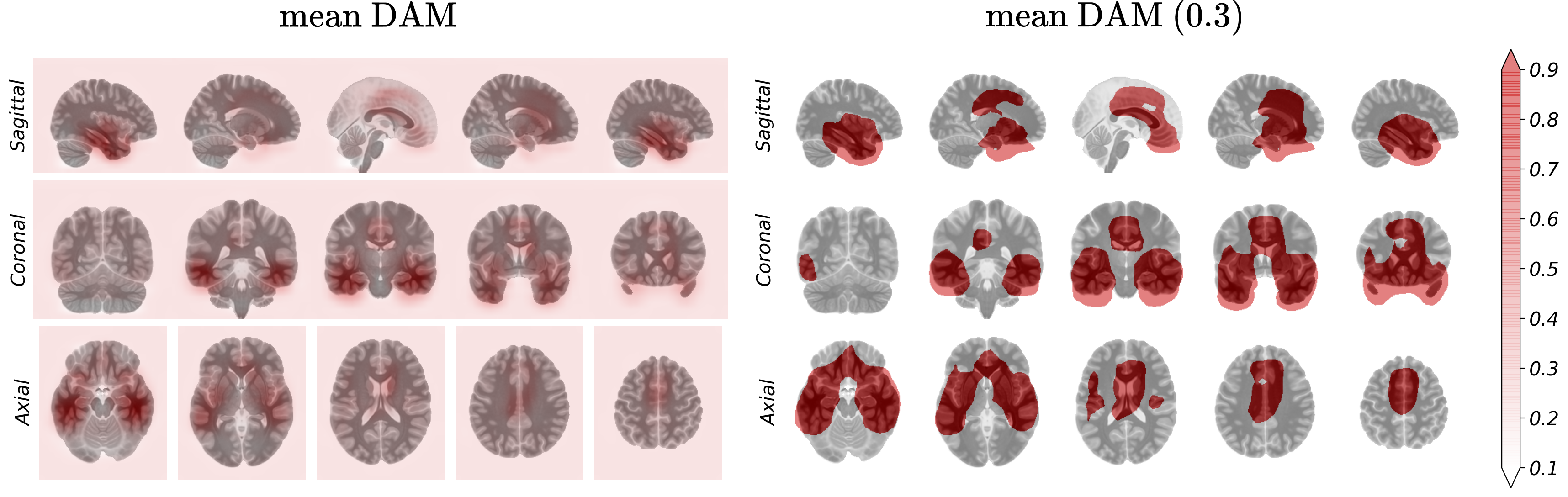}
\caption{Examples of predetermined discriminative brain regions inspired by \citep{lian2020attention}.} 
\label{fig:fig_7_1}
\end{subfigure}
\begin{subfigure}[b]{0.4\textwidth}
\centering
\includegraphics[width=\textwidth]{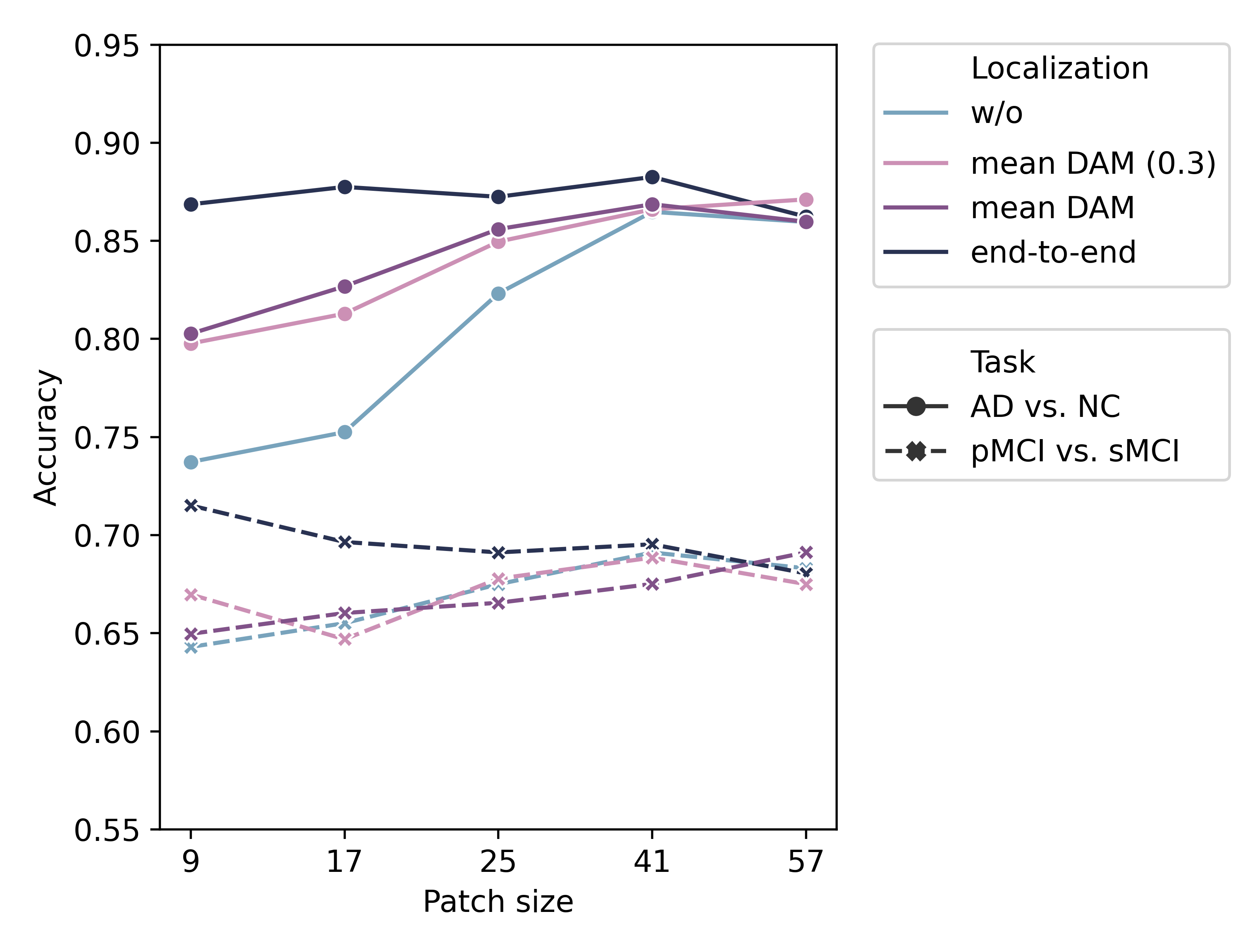}
\caption{Accuracy} 
\label{fig:fig_7_2}
\end{subfigure}
\begin{subfigure}[b]{0.4\textwidth}
\centering
\includegraphics[width=\textwidth]{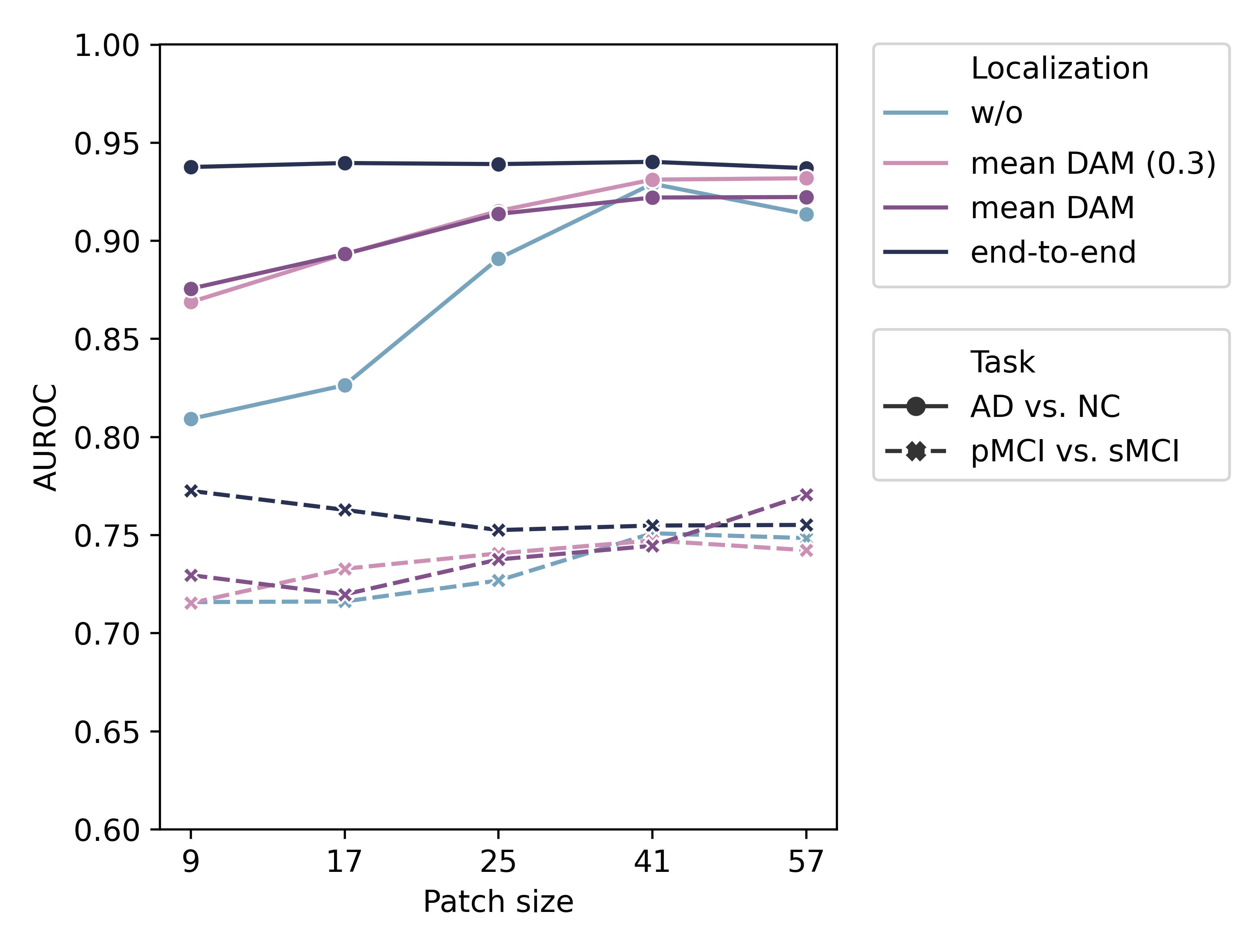}
\caption{AUROC} 
\label{fig:fig_7_3}
\end{subfigure}

\caption{Illustration of classification performance in terms of accuracy and area under the receiver operating characteristic (AUROC) by the patch size used in model training, localization method, and task.}
\label{fig:fig_7}
\end{figure}

\subsection{Ablation Study for Localization Methods}
We performed an ablation study of localization methods using our proposed framework to evaluate the effectiveness of joint learning of discriminative brain-region localization and disease identification.
We compared four localization methods: \quotes{w/o}, \quotes{mean DAM}, \quotes{mean DAM (0.3)}, and \quotes{end-to-end}.
The \quotes{w/o} denotes BrainBagNets, and the \quotes{end-to-end} denotes the proposed models, which are PG-BrainBagNets. Both \quotes{mean DAM} and \quotes{mean DAM (0.3)} were models trained with the predetermined discriminative brain region inspired by the mean DAM introduced in \citep{lian2020attention}.
For \quotes{mean DAM}, the proposed framework has been trained using predetermined $\mathbf{G}$ by considering the mean DAM to be $\mathbf{G}$ instead of training position embedding and gate network.
The resulting model was denoted as \quotes{mean DAM}.
In addition, we generated a binary mask because the representation was not used as a probabilistic value but instead was used for extracting patches in \citep{lian2020attention}.
In patch extraction, the threshold of 0.3 was used in the literature to represent potential patch locations.
We obtained a binary mask based on this threshold, and model training was performed in the same way as for \quotes{mean DAM}.
The predetermined $\mathbf{G}$ for the \quotes{mean DAM} and \quotes{mean DAM (0.3)} are illustrated in Fig. \ref{fig:fig_7_1}.

The average accuracy and AUROC in five-fold cross-validation are described in Figs. \ref{fig:fig_7_2} and \ref{fig:fig_7_3}.
First, when the model was trained without brain-region localization, classification performance decreased as the patch size reduced.
The model trained using the smallest patches exhibited the lowest classification performance for both tasks in accuracy and AUROC.
By adding the predetermined localization method, the classification performance improved compared with that without the localization method.
However, localization was performed regardless of the diagnosis model, resulting in a worse classification than the proposed method.
In the MCI conversion prediction task, only the proposed method demonstrated increased classification performance by limiting the increase in patch size.
This result implies that the regularization of the patch size allows extracting AD-related local and subtle changes but requires suitable brain-region localization dependent on the diagnosis model.

\begin{figure}[tb]
\centering    
\includegraphics[width=0.8\textwidth]{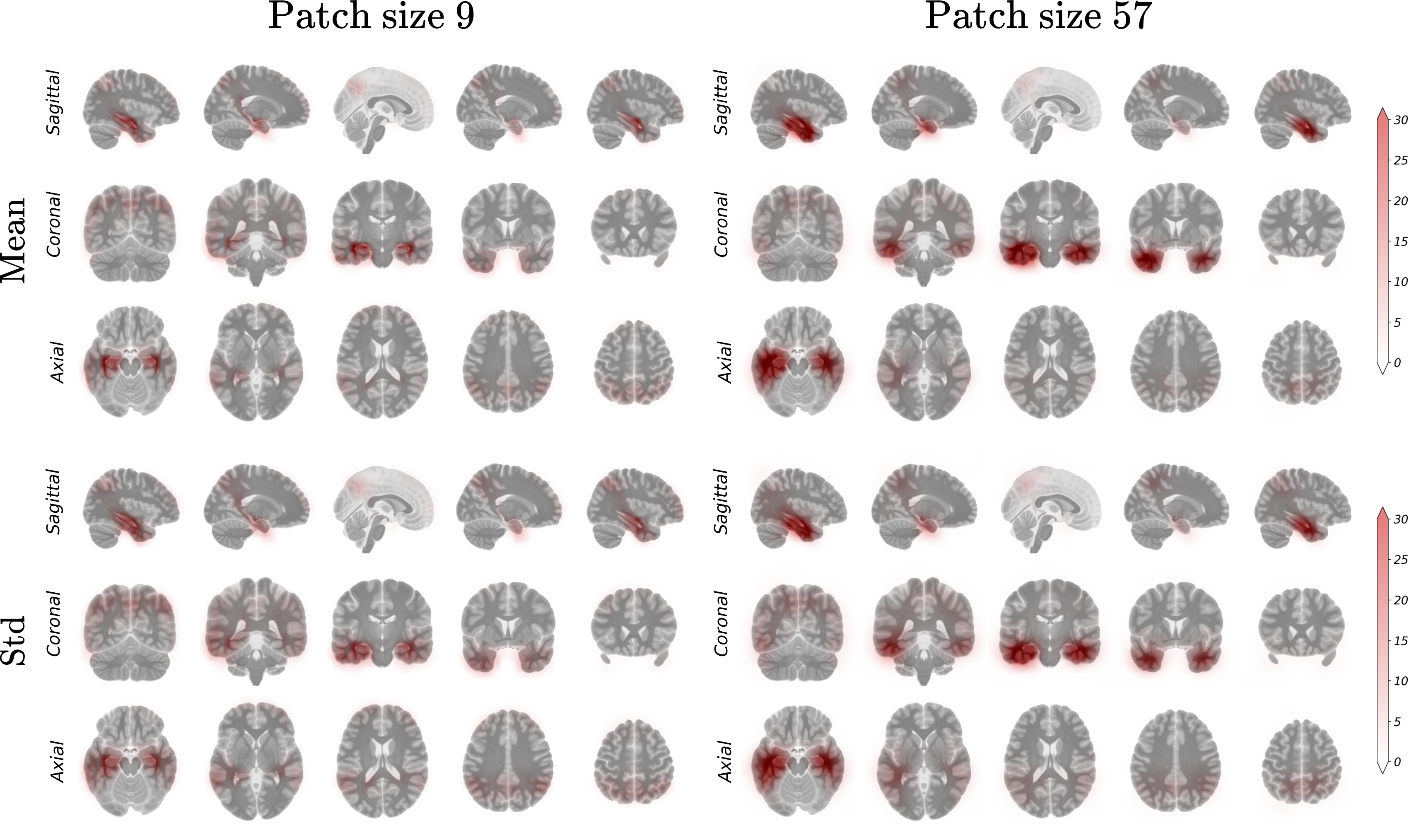}
\caption{Statistics of the positive patch-level class evidence drawn from true-positive examples.}
\label{fig:fig_8}
\end{figure}

\subsection{Regions of Disease Progression}
The proposed method was constructed by transparently aggregating patch-level responses for image-level predictions.
Thus, individualized AD progression regions were detected in a weakly supervised manner.
We confirmed patch-level responses over the testing dataset to analyze the detected regions where AD progression occurs.
For correctly predicted AD samples, patch-level class evidence was obtained, and positive values were left to obtain class evidence for AD class.
The results were aggregated over the testing dataset in five-fold cross-validation.
We calculated the point-wise mean and standard deviation, and the result were upsampled as the size of the MNI template, visualized in Fig. \ref{fig:fig_8}.

We observed that the high mean values were located in the hippocampal regions regardless of the patch size in the model. 
Additionally, the area where high standard deviation values were distributed was similar to that where the high mean values were distributed.
The high responses described in the standard deviation map revealed the high region variance where high patch-level class evidence appeared even across AD subjects.

\begin{figure}[tb]
\centering
\includegraphics[width=0.8\textwidth]{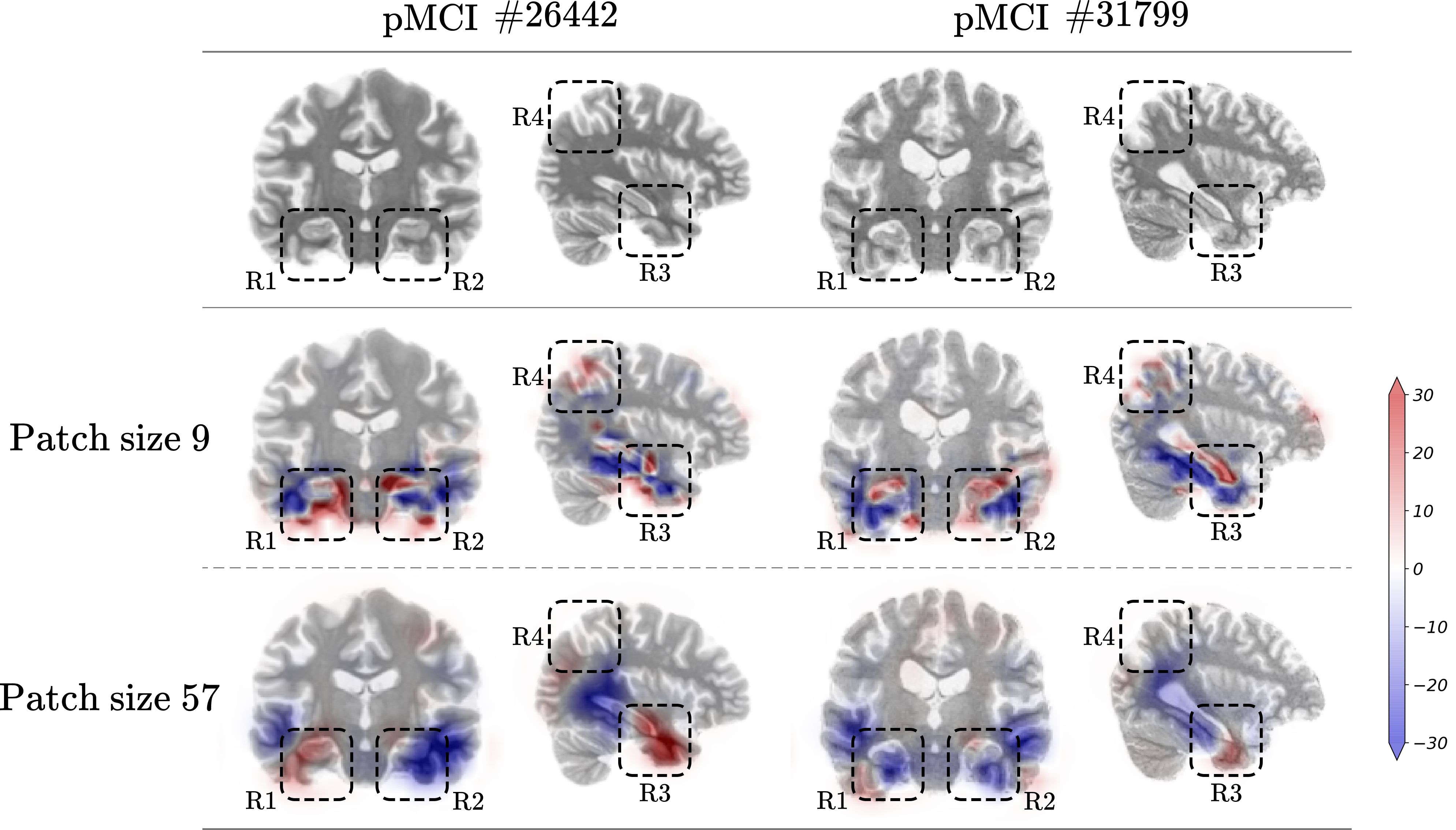}
\caption{Examples of false negatives from the model trained with large patches but correctly predicted by the model trained using small patches. Each column indicates one sample labeled as progressive mild cognitive impairment (pMCI), where the number next to the \# denotes the corresponding image ID of an input MRI scan. In addition, the blue and red colors indicate high class evidence for the stable mild cognitive impairment and pMCI class in that region, respectively.}
\label{fig:fig_9}
\end{figure}

\subsection{Effectiveness of Subtle Changes Captured using Small Patches}
We demonstrated that the model extracting the local class evidence captured using a small receptive field size better predicts MCI conversion.
We analyzed the two pMCI samples that yielded false negatives from the model trained using a large patch and made predictions correctly by limiting the patch size.
The local class evidence according to patch size is described in Fig. \ref{fig:fig_9}.
The first row depicted the original sMRI scans with image ID 26442 and 31799.
The following rows demonstrated the patch-level class evidence according to the patch size.
For a better comparison, the areas where a high amount of class evidence was contained were marked with dashed rectangles and denoted as R1 to R4.
The blue and red colors indicate high class evidence for sMCI and pMCI classes in that region, respectively.
The model trained using small patches displayed fine class evidence detected in local regions, whereas the class evidence found in models using large patches exhibited coarse patch-level class evidence results.

First, in the bottom of the R1 region, we observed that Sample \#26442 contains brain atrophy in the temporal lobe rather than the hippocampus, compared to Sample \#31799.
In contrast, Sample \#31799 depicts brain atrophies located in the hippocampal area.
These brain atrophies were correctly captured by the model trained using small patches.
However, the estimation produced by the model trained using larger patches demonstrated the difficulties of capturing these subtle changes.
These patterns can be observed in the R1, R2, and R3 regions. 
The positive class evidence for the pMCI class located in the parietal lobe area was only captured by the model trained using small patches, which can be observed in comparing the R4 region.
Finally, the model trained using small patches could determine sufficient local evidence to correctly predict the MCI conversion, whereas models trained using large patches could not capture sufficient cues for a correct decision.
In this analysis, we observed that regularizing the increasing patch size increased the prediction performance for MCI conversion by extracting subtle and local structural feature representation.


\section{Conclusion}
The sMRI-based deep learning approaches have been widely used and continue evolving. 
Many studies have focused on subtle brain atrophy to better understand AD biomarkers in sMRI. 
As not all subtle structural changes are associated with brain disease, discriminative brain-region localization has attracted attention.
To alleviate the problem of brain-region localization for patch-level feature representation, we proposed a framework for jointly learning discriminative brain-region localization and disease identification in an end-to-end manner.
In the experiment, we evaluated the proposed method on the ADNI dataset. 
The proposed method displayed the best classification performance in both the AD diagnosis and MCI conversion prediction task compared with existing methods. 
Primarily, the proposed method effectively increased the classification performance, when localization of the subtle changes was required.
We also demonstrated the interpretability of the proposed method by tracing the rationale for the model predictions down to the small patch level.

\section*{Acknowledgement} 
This work was supported by Institute of Information \& communications Technology Planning \& Evaluation (IITP) grant funded by the Korea government (MSIT) under Grant 2019-0-00079 (Department of Artificial Intelligence(Korea University)) and under Grant 2017-0-01779 (a machine learning and statistical inference framework for explainable
artificial intelligence).

\bibliographystyle{elsarticle-harv}

\bibliography{main}



\end{document}